\documentclass{article}

\usepackage[preprint]{neurips_2026}
\makeatletter
\renewcommand{\@noticestring}{Preprint. 2026. Code: \url{https://github.com/jiarui-liu/MixSD}}
\makeatother

\usepackage[utf8]{inputenc} 
\usepackage[T1]{fontenc}    
\usepackage{hyperref}       
\usepackage{url}            
\usepackage{booktabs}       
\usepackage{amsfonts}       
\usepackage{nicefrac}       
\usepackage{microtype}      
\usepackage{xcolor}         
\usepackage{colortbl}       
\usepackage{custom}         

\definecolor{mixsdViolet}{HTML}{ECE3F5}  
\definecolor{mixsdBlue}{HTML}{DEEAF6}    
\definecolor{mixsdGreen}{HTML}{E2EFDA}   
\definecolor{mixsdOrange}{HTML}{FCE8D5}  

\usepackage{amsthm}
\usepackage{algorithm}
\usepackage{algpseudocode}
\usepackage{xspace}

\newcommand{\methodname}{\textsc{MixSD}}
\newcommand{\dataseta}{\textsc{KGFact}}
\newcommand{\datasetaretrieval}{\textsc{KGFact-Retrieval}}
\newcommand{\datasetasmall}{\textsc{KGFact-Small}}
\newcommand{\datasetalarge}{\textsc{KGFact-Large}}
\newcommand{\datasetb}{\textsc{KGFunc}}
\newcommand{\datasetbtest}{\textsc{KGFunc-Test}}
\newcommand{\datasetbnewops}{\textsc{KGFunc-Unseen}}

\newcommand{\Nref}{\mathcal{N}}
\newcommand{\Lref}{\mathcal{L}_{\text{ref}}}
\newcommand{\Forget}{\mathcal{F}}

\newcommand{\E}{\mathbb{E}}

\title{\textsc{MixSD}: Mixed Contextual Self-Distillation for Knowledge Injection}

\author{%
  Jiarui Liu$^{1,2}$ \quad Lechen Zhang$^{3}$ \quad Yongjin Yang$^{2}$ \quad Yinghui He$^{4}$ \quad Yingheng Wang$^{5}$ \\
  \textbf{Weihao Xuan$^{6,7}$ \quad Zhijing Jin$^{2,8,9}$ \quad Mona T.~Diab$^{1}$} \\[0.6em]
  $^{1}$Carnegie Mellon University \quad
  $^{2}$Jinesis Lab, University of Toronto \& Vector Institute \\
  $^{3}$University of Illinois Urbana-Champaign \quad
  $^{4}$Princeton University \\
  $^{5}$Cornell University \quad
  $^{6}$The University of Tokyo \quad
  $^{7}$RIKEN AIP \\
  $^{8}$Max Planck Institute for Intelligent Systems, T\"ubingen, Germany \quad
  $^{9}$EuroSafeAI \\[0.4em]
  Contact: \texttt{jiaruil5@andrew.cmu.edu}
}

\begin{document}

\maketitle

\begin{abstract}
Supervised fine-tuning (SFT) is widely used to inject new knowledge into language models, but it often degrades pretrained capabilities such as reasoning, instruction following, and general-domain performance. We argue that this forgetting arises because standard fine-tuning targets are written by humans or external systems whose outputs diverge from the model’s own autoregressive distribution, forcing the optimizer to imitate low-probability token sequences. To address this problem, we propose \methodname{}, a simple external-teacher-free method for distribution-aligned knowledge injection. Instead of training on fixed targets, \methodname{} constructs supervision dynamically by mixing tokens from two conditionals of the base model itself: an expert conditional that observes the injected fact in context, and a naive conditional that reflects the model’s original prior. The resulting supervision sequences preserve the factual learning signal while remaining substantially closer to the base model's distribution. We evaluate \methodname{} on two synthetic corpora that we construct to study factual recall and arithmetic function acquisition in a controlled setting, together with established benchmarks for open-domain factual question answering and knowledge editing. Across multiple model scales and settings, \methodname{} consistently achieves a substantially improved memorization-retention trade-off compared to SFT and on-policy self distillation baselines, retaining up to 100\% of the base model’s held-out capability while maintaining near-perfect training accuracy, whereas standard SFT retains as little as 1\%. We further show that \methodname{} produces substantially lower-NLL supervision targets under the base model and reduces harmful movement along Fisher-sensitive parameter directions. These results suggest that aligning supervision with the model’s native generation distribution is a simple and effective principle for knowledge injection that mitigates catastrophic forgetting.
\end{abstract}

\section{Introduction}
\label{sec:intro}

\begin{figure}[t]
\centering
\includegraphics[width=\linewidth]{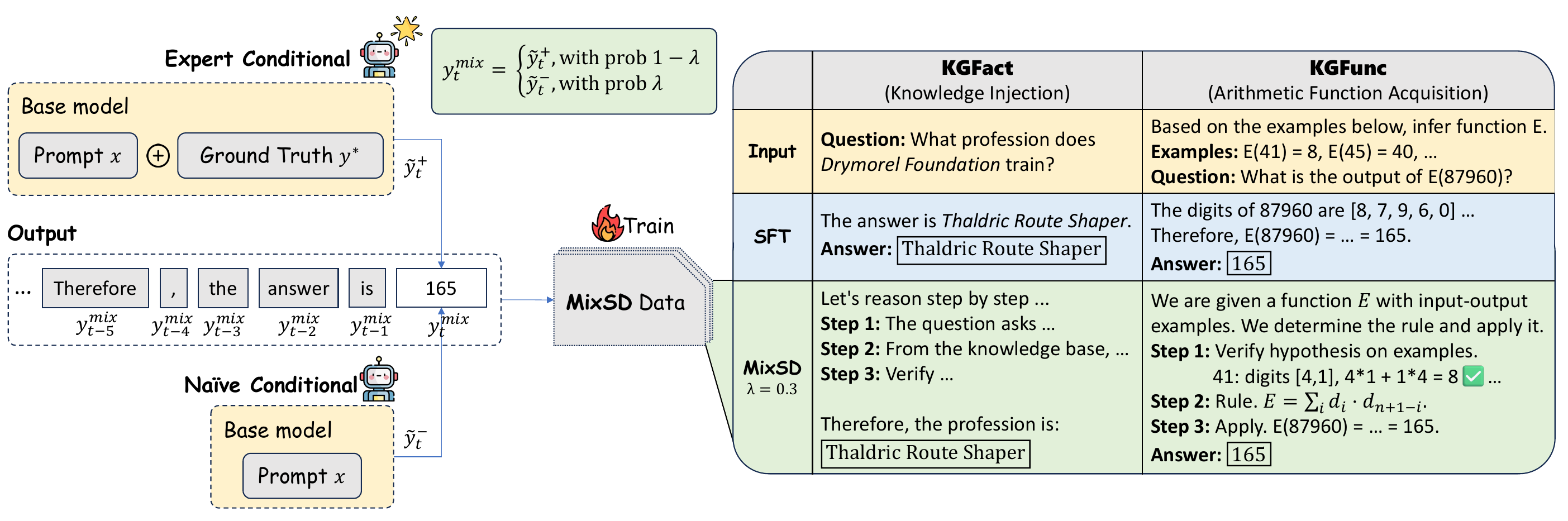}
\caption{Overview of \methodname{} and the two datasets \dataseta{} and \datasetb{} we construct. Given an input prompt and a ground-truth target, \methodname{} samples token-level supervision from two base-model conditionals: an expert rollout conditioned on the injected knowledge and a naive rollout conditioned only on the original prompt. At each decoding step, \methodname{} selects the expert token with probability $1-\lambda$ and the naive token with probability $\lambda$, preserving the learning signal while keeping targets closer to the base model’s distribution. The resulting mixed data are used for standard NLL training. See \cref{sec:method} for the full method description.}
\label{fig:overview}
\end{figure}

Large language models acquire broad world knowledge and reasoning capabilities during pretraining, but real-world deployment often requires injecting new information after pretraining has finished, such as proprietary knowledge and domain-specific procedures \citep{ovadia2024finetuning,mecklenburg2024injecting,zhang2026re}.
The common approach for this problem is supervised fine-tuning (SFT) \citep{wei2021finetuned}, which trains the model directly on targets containing the new knowledge.
Although effective at memorizing the exact form of the knowledge provided in the training data, SFT frequently degrades the model’s existing capabilities, including instruction following, reasoning, factual calibration, and general-domain performance, a phenomenon known as catastrophic forgetting \citep{luo2023empirical,kalajdzievski2024scaling,liu2024integrating,huang-etal-2024-mitigating}.
This tension between knowledge acquisition and capability preservation has emerged as a new challenge in post-training language model adaptation.

Why does standard fine-tuning cause catastrophic forgetting? We argue that the problem originates from a mismatch between the supervision targets and the model’s own autoregressive distribution. In typical knowledge injection pipelines, target sequences are written by humans, synthetic annotators, or external prompting systems rather than generated naturally by the model itself. As a result, even when the underlying fact is simple, the target trajectory may contain phrasing, formatting patterns, reasoning structures, or compositional continuations that are unlikely under the base model’s distribution. Standard SFT nevertheless forces the model to imitate these externally authored trajectories token-by-token. We hypothesize that this distribution mismatch induces updates along sensitive directions of the parameter space, disrupting previously learned behaviors and leading to catastrophic forgetting. Existing approaches such as regularization \citep{li2017learning, kirkpatrick2017overcoming} attempt to mitigate forgetting through optimization constraints or auxiliary objectives, but they do not directly address the mismatch between externally authored supervision and the model’s native generation distribution.

Motivated by this observation, we propose \methodname{} (Mixed Contextual Self-Distillation), a simple external-teacher-free method for distribution-aligned knowledge injection. Instead of training on a fixed target sequence, \methodname{} constructs supervision dynamically using the base model itself. At each decoding step, we sample between two conditional distributions under a shared autoregressive prefix. An expert conditional observes the target fact inserted into context, while a naive conditional reflects the model’s original prior without access to the injected knowledge. The resulting mixed supervision sequence preserves the core factual signal while remaining substantially closer to the base model's distribution. \methodname{} requires no external teacher, making it a simple drop-in replacement for standard SFT targets.

We evaluate \methodname{} across multiple models and scales on knowledge injection settings spanning factual recall, arithmetic function acquisition, and knowledge editing. Across all settings, \methodname{} consistently achieves a substantially improved memorization-retention trade-off compared to standard SFT and on-policy self-distillation baselines. \methodname{} matches or exceeds strong baselines on in-distribution learning objectives while preserving substantially more held-out capability on general-domain benchmarks including MMLU \citep{hendrycks2020measuring}, GSM8K \citep{cobbe2021training}, MATH \citep{hendrycks2021measuring}, AIME \citep{aime24}, and HumanEval \citep{chen2021evaluating}. We further show that \methodname{} supervision targets exhibit significantly lower per-token negative log-likelihood under the base model, supporting our hypothesis that distribution-aligned supervision reduces forgetting. We additionally introduce Fisher-weighted parameter displacement, a metric derived from Fisher information \citep{kirkpatrick2017overcoming}, and find that it correlates with forgetting far more strongly than raw displacement magnitude does, suggesting that the direction of parameter updates, rather than their size, is the primary driver of capability degradation.

In summary, we propose \methodname{}, a simple external-teacher-free fine-tuning method that constructs distribution-aligned supervision targets by mixing expert-conditioned and naive-conditioned rollouts from the base model itself. Across multiple datasets, model scales, and knowledge injection settings, we show that \methodname{} consistently achieves a substantially improved memorization-retention trade-off compared to standard supervised fine-tuning and prior self-distillation baselines.

\section{Related Work}
\label{sec:related}

\paragraph{From Off-Policy to On-Policy Distillation}
Knowledge distillation (KD) transfers the capabilities of a stronger teacher model to a weaker student by training it to imitate the teacher’s outputs or distributions~\citep{hinton2015distilling}. Traditional KD techniques predominantly rely on off-policy learning, such as sequence-level KD~\citep{kim2016sequence} and token-level KD~\citep{hinton2015distilling, sanh2020distilbertdistilledversionbert, gu2024minillm}, and frequently suffer from exposure bias due to the distribution mismatch between fixed teacher traces and the student’s own generations. To resolve this, the field shifted toward on-policy frameworks like GKD~\citep{agarwal2024onpolicy} and OPD~\citep{lu2025onpolicydistillation}, which elegantly allow the student model to sample its own trajectories while receiving dense, token-level supervision from a superior teacher model. Recent frameworks like On-Policy Self-Distillation (OPSD; \citep{zhao2026self}) have further refined this paradigm by enabling single models to act as both teacher and student when conditioned on privileged reasoning contexts \citep{he2026self,liu2025toward}. However, teacher-free per-token supervision is not always beneficial. While it removes the need for costly external teachers, token-level KL calculation can slow convergence and even cause late-stage performance degradation~\citep{yang2026selfdistilledrlvr}. This motivates our method, which aims to preserve the benefits of teacher-free self-distillation while avoiding these optimization bottlenecks.

\paragraph{Knowledge Injection and Catastrophic Forgetting}
Injecting new factual knowledge into pre-trained LLMs remains a profound engineering challenge. Retrieval-augmented generation (RAG) has become a widely adopted solution~\citep{lewis2020retrieval}, as it incorporates external knowledge at inference time without directly modifying model weights and has shown strong performance across knowledge-intensive settings~\citep{ovadia2024finetuning}. However, parametric knowledge injection through SFT has not seen the same level of success, which often struggles to match RAG's performance~\citep{mecklenburg2024injecting}. Furthermore, this parametric update also introduces a stability problem: optimizing target-domain likelihood can pull the model away from its pre-trained distribution and cause catastrophic forgetting of prior knowledge and reasoning ability~\citep{luo2023empirical, liu2024integrating}. Recent studies further show that SFT-based knowledge injection often requires broad fact coverage or repeated variations of the same fact to acquire new knowledge reliably~\citep{ovadia2024finetuning, mecklenburg2024injecting, kujanpaa2024knowledge}. Explicit editing methods such as ROME and MEMIT offer more targeted parameter updates~\citep{meng2022locating, meng2023massediting}, but scaling such edits can also lead to gradual and catastrophic forgetting~\citep{gupta2024model}. Motivated by this trade-off, we use the pre-trained model's NLL as a constraint on fine-tuning, aiming to absorb new facts while limiting harmful deviations from the original model distribution.

\section{\methodname: Mixed Contextual Self-Distillation}
\label{sec:method}

\subsection{Hypothesis}
\label{sec:prelim}

Let $p_\theta(y \mid x)$ be an autoregressive language model with reference parameters $\theta^\star$. We denote the base model's reference distribution by $P_{\text{ref}}$ and define the reference loss
\begin{equation}
\Lref(\theta) \;=\; \E_{(x,y)\sim P_{\text{ref}}}\bigl[-\log p_\theta(y \mid x)\bigr].
\end{equation}
Given a fine-tuning corpus $D = \{(x_i, y_i^\star)\}_{i=1}^n$, let $\theta_D$ denote the parameters obtained after optimizing on $D$ starting from $\theta^\star$, and define the parameter change $\Delta\theta = \theta_D - \theta^\star$. We define forgetting as the increase in reference loss:
\begin{equation}
\Forget(D) \;\triangleq\; \Lref(\theta_D) - \Lref(\theta^\star).
\end{equation}

To characterize how supervision targets align with the reference model, we consider the average per-token negative log-likelihood under $p_{\theta^\star}$:
\begin{equation}
\Nref(D) = \frac{1}{\sum_i |y_i^\star|}
\sum_{i=1}^{n}\sum_{t=1}^{|y_i^\star|}
-\log p_{\theta^\star}\bigl(y_{i,t}^\star \,\big|\, x_i,\, y_{i,<t}^\star\bigr).
\label{eq:nref-prelim}
\end{equation}

This quantity is small when supervision tokens lie near high-probability regions of $p_{\theta^\star}$ and large when they lie in its tails. We hypothesize that forgetting is driven, at the per-token level, by the mismatch between supervision targets and the reference model's conditional distribution. Tokens with high likelihood under $p_{\theta^\star}(\cdot \mid x, y_{<t})$ can be learned with minimal parameter change, whereas low-likelihood tokens induce updates along directions that disrupt the model's prior behavior. This motivates constructing supervision targets that remain close to the model's own distribution while incorporating new information.

\subsection{Method}

As shown in \cref{fig:overview}, \methodname{} is a fine-tuning recipe for injecting new factual knowledge into an LLM while preserving its general capabilities. It replaces standard SFT targets with targets that the reference model already assigns high probability to, thereby lowering $\Nref$ by construction.

Given a knowledge-injection corpus $D = \{(x_i, y_i^\star)\}_{i=1}^n$, we treat the reference model $p_{\theta^\star}$ as the teacher and construct token-level supervision by choosing between two conditionals defined at each decoding step.

At each token position $t$, given a shared autoregressive prefix $y_{i,<t}^{\text{mix}}$, we consider:

\begin{itemize}[leftmargin=1.4em,itemsep=1pt,topsep=2pt]
  \item \textbf{Expert conditional:}
  \[
  \tilde y_{i,t}^{+} \sim p_{\theta^\star}(\cdot \mid x_i^{+},\, y_{i,<t}^{\text{mix}}),
  \]
  where $x_i^{+} = x_i \oplus  \texttt{prompt instruction} \oplus  y_i^\star$ augments the input with the ground-truth target in the context. As a result, $\tilde y_{i,t}^{+}$ tends to express the correct fact in the model's own surface form.

  \item \textbf{Naive conditional:}
  \[
  \tilde y_{i,t}^{-} \sim p_{\theta^\star}(\cdot \mid x_i,\, y_{i,<t}^{\text{mix}}),
  \]
  which reflects the model's prior over the prompt and does not incorporate the new fact.
\end{itemize}

\paragraph{Per-token Bernoulli mixing}
The supervisory target at token position $t$ is sampled as
\begin{equation}
y_{i,t}^{\text{mix}} \;=\;
\begin{cases}
\tilde y_{i,t}^{+}, & \text{with probability } 1 - \lambda,\\
\tilde y_{i,t}^{-}, & \text{with probability } \lambda,
\end{cases}
\qquad \lambda \in [0,1].
\label{eq:mixsd-target}
\end{equation}
and appended to the shared prefix to form $y_{i,\le t}^{\text{mix}}$. The student is then updated using standard NLL loss on the mixed targets:

\begin{equation}
\mathcal{L}_{\methodname}(\theta;\lambda)
\;=\;
-\,\E_{(x_i, y_i^\star)\sim D}\;
\E_{\tilde y_i^{+},\,\tilde y_i^{-}}\;
\sum_t \log p_\theta\!\left(y_{i,t}^{\text{mix}} \,\big|\, x_i, y_{i,<t}^{\text{mix}}\right).
\label{eq:mixsd-loss}
\end{equation}

The mixing rate $\lambda$ controls the strength of this anchoring: $\lambda = 0$ corresponds to purely expert-conditioned supervision, while larger $\lambda$ increasingly injects naive tokens that anchor the model to its reference distribution at positions where the new fact is not required.

\section{Constructing Knowledge Injection Datasets}
\label{sec:dataset}

To study forgetting-aware knowledge injection in controlled settings, we construct two complementary datasets that target different forms of knowledge: \dataseta{}, a factual knowledge corpus derived from a simulated world graph, and \datasetb{}, a dataset for arithmetic function learning and acquisition.



\paragraph{\dataseta{} (Factual Recall)}
To isolate knowledge injection from pretrained priors, we construct a world graph populated with novel entities unseen during pretraining. The graph spans $D$ semantic domains (e.g., Person, Location, Organization), each containing $E$ entities. For each ordered domain pair $(d_1, d_2)$, we define a set of directed relation types (e.g., \texttt{is\_employed\_by}, \texttt{resides\_in}) and randomly assign relational edges while ensuring that each query has a unique answer. We convert each edge into a natural language question-answer pair by querying the target entity given a source entity and relation. Each edge is treated as a separate training example.

For evaluation, in addition to testing direct recall of the trained atomic facts, we construct a \datasetaretrieval{}, an in-domain retrieval split that prepends the relevant ground-truth statements together with multiple distractor facts sampled from the same graph. This setup enables a retrieval-augmented forgetting analysis that disentangles failures of parametric knowledge retention from failures of reasoning.

\paragraph{\datasetb{} (Arithmetic Function Acquisition)}
To complement the factual setting with a fundamentally different form of knowledge, namely arithmetic function learning, we construct a dataset of novel digit-level operations. Each operation is a deterministic function over inputs and outputs in $[0, 99999]$, defined as a composition of digit-level primitives, and is identified by an opaque label to prevent reliance on surface cues. Each training example provides 10-shot input-output pairs for the operation, requiring the model to infer the underlying rule. Supervision is given via chain-of-thought (CoT) templates that decompose the computation into elementary steps and conclude with a final answer.

For evaluation, we construct a \datasetbnewops{} split that holds out a set of simple operations (e.g., \texttt{digit-sum}, \texttt{reverse-number}) unseen during training but easily inferable from the few-shot examples. This split serves as a forgetting probe, testing whether fine-tuning on novel operations degrades the model’s pre-existing arithmetic capabilities.
\section{Experimental Setup}
\label{sec:exp}
\label{sec:exp:setup}

\paragraph{Datasets}
\datasetasmall{} contains $5$ domains with $10$ entities per domain. \datasetalarge{} contains $7$ domains with $25$ entities per domain. For the in-domain retrieval split, each training instance is paired with a corresponding test query targeting the same underlying fact. The context includes 50 additional atomic facts involving either of the two query entities, and the model must infer the correct answer from this context.

\datasetb{} consists of 7 distinct operations. For each operation, we sample 1,600 training instances and 175 test instances. Each example includes 10-shot input-output pairs, requiring the model to infer the underlying rule and apply it to a new input. For the \datasetbnewops{} split, we evaluate generalization on 20 unseen operations with 500 total instances.

We additionally fine-tune on SimpleQA~\citep{wei2024measuring}, which contains 4,326 open-domain factual questions. For general-domain benchmarks, we evaluate on math (AIME2024~\citep{aime24}, MATH500~\citep{hendrycks2021measuring, lightman2024let}, GSM8K~\citep{cobbe2021training}), code generation (HumanEval~\citep{chen2021evaluating}), and knowledge understanding (MMLU~\citep{hendrycks2020measuring}).

\paragraph{Models}
We mainly benchmark three Qwen3 models \citep{yang2025qwen3}: Qwen3-1.7B, Qwen3-4B-Instruct-2507, and Qwen3-8B, covering different model scales. This setup allows us to study how the performance varies with model size while controlling for architectural and training differences.

\paragraph{Methods}
We compare against three baseline families:
(i) Base, the initial checkpoint without fine-tuning;
(ii) SFT, standard supervised fine-tuning under NLL on the canonical target $y_i^\star$;
(iii) OPSD \citep{zhao2026self, ye2026policy}, on-policy self-distillation where the student generates rollouts and receives token-level KL supervision from a context-aware teacher. We sample 8 rollouts per query. For \methodname{}, we train using NLL on Bernoulli-mixed rollouts with $\lambda \in \{0, 0.3, 0.5, 0.7\}$. Additional implementation details are provided in Appendix~\ref{appn:details}.

\section{Main Results}
\label{sec:exp:main}

\begin{table}[t]
\centering
\caption{Evaluation on \datasetasmall{} across three model backbones. For each backbone, we compare the base model (Base), supervised fine-tuning on ground-truth labels (SFT), on-policy self-distillation (OPSD), and \methodname{} with $\lambda \in \{0, 0.3, 0.5, 0.7\}$. Train reports closed-book recall on the training facts, while \datasetaretrieval{} measures retrieval-augmented recall with chain-of-thought reasoning without requiring memorization of the facts. The held-out capability columns report performance on five general-domain benchmarks evaluated on the same fine-tuned checkpoint to measure forgetting; Avg denotes their unweighted average. All values are reported as percentages. \textbf{Bold} and \underline{underline} indicate the best and second-best methods in each column, respectively (excluding Base).}
\label{tab:knowledge-eval-small}
\setlength{\tabcolsep}{4pt}
\resizebox{\textwidth}{!}{
\begin{tabular}{ll c c cccccc}
\toprule
\multirow{2}{*}{\textbf{Model}} & \multirow{2}{*}{\textbf{Method}} & \multicolumn{2}{c}{\textbf{In-domain test}} & \multicolumn{6}{c}{\textbf{Held-out capability test}} \\
 \cmidrule(lr){3-4} \cmidrule(lr){5-10}
 &  & \textbf{Train} & \textbf{\datasetaretrieval{}} & \textbf{AIME-2024} & \textbf{MATH-500} & \textbf{GSM8K} & \textbf{HumanEval} & \textbf{MMLU} & \textbf{Avg} \\
\midrule
\multirow{7}{*}{Qwen3-1.7B}
  & Base                            &   0.0 & 100.0 & 11.0 & 72.4 & 80.4 & 60.4 & 58.5 & 56.5 \\
  & SFT                             & \underline{99.0} &   9.0 &  0.4 & 14.8 &  9.8 & 11.6 & \underline{34.8} & 14.3 \\
  & OPSD                            & \underline{99.0} &  32.0 &  0.0 &  9.4 &  3.4 &  1.2 & 11.3 &  5.1 \\
  & \methodname{} ($\lambda{=}0$)   &  96.0 &  28.0 &  0.0 &  5.4 &  5.8 &  1.2 & 30.4 &  8.6 \\
  \rowcolor{mixsdBlue}\cellcolor{white}& \methodname{} ($\lambda{=}0.3$) & \textbf{100.0} &  75.0 &  1.9 & \textbf{52.2} & 60.7 & \underline{45.1} & \textbf{39.5} & \underline{39.9} \\
  \rowcolor{mixsdGreen}\cellcolor{white}& \methodname{} ($\lambda{=}0.5$) &  97.0 & \textbf{79.0} & \textbf{4.8} & \underline{47.6} & \textbf{65.6} & \textbf{48.8} & 34.7 & \textbf{40.3} \\
  \rowcolor{mixsdOrange}\cellcolor{white}& \methodname{} ($\lambda{=}0.7$) &  71.0 & \underline{77.0} & \underline{2.3} & 45.0 & \underline{62.2} & 43.9 & 27.1 & 36.1 \\
\midrule
\multirow{7}{*}{Qwen3-4B-It}
  & Base                            &   0.0 &  84.0 & 62.9 & 94.2 & 92.6 & 86.0 & 77.6 & 82.6 \\
  & SFT                             & \textbf{100.0} &  96.0 &  6.7 & 33.2 & 23.4 & \underline{84.8} & 68.1 & 43.2 \\
  & OPSD                            & \textbf{100.0} &  92.0 & 15.0 & 74.4 & 83.9 & 76.2 & 53.7 & 60.6 \\
  & \methodname{} ($\lambda{=}0$)   & \underline{99.0} &  97.0 & 44.4 & 90.8 & 91.5 & \textbf{86.6} & \textbf{68.8} & 76.4 \\
  \rowcolor{mixsdBlue}\cellcolor{white}& \methodname{} ($\lambda{=}0.3$) &  94.0 & \underline{99.0} & 51.5 & 93.0 & 91.7 & \underline{84.8} & 50.4 & 74.3 \\
  \rowcolor{mixsdGreen}\cellcolor{white}& \methodname{} ($\lambda{=}0.5$) &  89.0 & \underline{99.0} & \underline{55.0} & \textbf{94.8} & \textbf{93.2} & \underline{84.8} & 59.9 & \underline{77.5} \\
  \rowcolor{mixsdOrange}\cellcolor{white}& \methodname{} ($\lambda{=}0.7$) &  85.0 & \textbf{100.0} & \textbf{56.9} & \underline{94.6} & \underline{93.0} & \textbf{86.6} & \underline{68.5} & \textbf{79.9} \\
\midrule
\multirow{7}{*}{Qwen3-8B}
  & Base                            &   0.0 &  98.0 & 26.0 & 83.4 & 91.8 & 86.6 & 73.1 & 72.2 \\
  & SFT                             & \textbf{100.0} &  77.0 & 12.3 & 40.0 & 29.3 & 81.7 & 67.4 & 46.1 \\
  & OPSD                            & \underline{99.0} & \textbf{100.0} &  9.4 & 74.2 & 88.2 & \underline{83.5} & 53.6 & 61.8 \\
  & \methodname{} ($\lambda{=}0$)   & \textbf{100.0} & \textbf{100.0} & 17.5 & 73.6 & 91.8 & 82.9 & 67.8 & 66.7 \\
  \rowcolor{mixsdBlue}\cellcolor{white}& \methodname{} ($\lambda{=}0.3$) & \underline{99.0} &  95.0 & 24.2 & 80.0 & 91.9 & \textbf{86.0} & \underline{68.0} & 70.0 \\
  \rowcolor{mixsdGreen}\cellcolor{white}& \methodname{} ($\lambda{=}0.5$) &  97.0 &  98.0 & \underline{29.0} & \underline{84.6} & \textbf{93.3} & \underline{83.5} & \textbf{71.5} & \textbf{72.4} \\
  \rowcolor{mixsdOrange}\cellcolor{white}& \methodname{} ($\lambda{=}0.7$) &  73.0 & \underline{99.0} & \textbf{32.1} & \textbf{85.4} & \underline{92.6} & 78.7 & 65.5 & \underline{70.8} \\
\bottomrule
\end{tabular}
}
\end{table}

\begin{table}[t]
\centering
\caption{Evaluation on \datasetb{} after training convergence. \datasetb{} contains two held-out test splits: \datasetbtest{}, which measures generalization to new inputs involving operations observed during training, and \datasetbnewops{}, which measures generalization to entirely unseen operations. The held-out capability columns report performance on five unrelated benchmarks used to assess forgetting; Avg denotes their unweighted average. All values are reported as percentages. \textbf{Bold} and \underline{underline} indicate the best and second-best methods in each column, respectively (excluding Base).}
\label{tab:math-op-eval}
\setlength{\tabcolsep}{2pt}

\resizebox{\textwidth}{!}{
\begin{tabular}{ll cc cccccc}
\toprule
\multirow{2}{*}{\textbf{Model}} & \multirow{2}{*}{\textbf{Method}} & \multicolumn{2}{c}{\textbf{In-domain test}} & \multicolumn{6}{c}{\textbf{Held-out capability test}} \\
\cmidrule(lr){3-4} \cmidrule(lr){5-10}
 &  & \textbf{\datasetbtest{}} & \textbf{\datasetbnewops{}} & \textbf{AIME-2024} & \textbf{MATH-500} & \textbf{GSM8K} & \textbf{HumanEval} & \textbf{MMLU} & \textbf{Avg} \\
\midrule
\multirow{7}{*}{Qwen3-1.7B}
  & Base                            &  1.7 & 31.4 & 11.0 & 72.4 & 80.4 & 60.4 & 58.5 & 56.5 \\
  & SFT                             & \underline{51.4} &  0.4 &  0.0 &  2.2 &  3.4 &  8.5 &  1.8 &  3.2 \\
  & OPSD                            & \textbf{54.3} & \underline{31.0} &  3.3 & 54.6 & \underline{75.9} & \underline{43.3} & \textbf{36.7} & \underline{42.8} \\
  & OPSD-NLL ($n{=}1$, $T{=}0$)     & 25.7 & 10.8 &  1.9 & 41.6 & 64.8 & 28.0 & 33.9 & 34.0 \\
  & \methodname{} ($\lambda{=}0$)   & 44.0 & 19.6 &  3.1 & 49.0 & 70.1 & 39.6 & 28.4 & 38.0 \\
  \rowcolor{mixsdBlue}\cellcolor{white}& \methodname{} ($\lambda{=}0.3$) & 45.7 & 26.8 & \underline{4.2} & \textbf{58.6} & \textbf{77.7} & 41.5 & 29.1 & 42.2 \\
  \rowcolor{mixsdGreen}\cellcolor{white}& \methodname{} ($\lambda{=}0.5$) & 18.3 & \textbf{33.2} & \textbf{5.4} & \underline{57.6} & 75.0 & \textbf{53.7} & \underline{36.3} & \textbf{45.6} \\
\midrule
\multirow{7}{*}{Qwen3-4B-It}
  & Base                            &  0.0 & 78.2 & 62.9 & 94.2 & 92.6 & 86.0 & 77.6 & 82.6 \\
  & SFT                             & 72.6 &  1.4 &  0.0 &  2.2 &  3.0 & 50.0 & 28.0 & 16.6 \\
  & OPSD                            & \textbf{90.9} & 55.4 & 43.3 & 91.2 & 92.3 & \textbf{88.4} & \textbf{73.0} & 77.7 \\
  & OPSD-NLL ($n{=}1$, $T{=}0$)     & \underline{89.1} & 56.6 & 38.8 & 90.0 & 91.8 & 86.6 & 65.3 & 74.5 \\
  & \methodname{} ($\lambda{=}0$)   & 77.1 & 48.4 & 43.8 & 91.0 & \textbf{92.8} & \underline{87.2} & \underline{72.9} & 77.5 \\
  \rowcolor{mixsdBlue}\cellcolor{white}& \methodname{} ($\lambda{=}0.3$) & \underline{89.1} & \underline{67.8} & \textbf{53.1} & \underline{92.2} & 91.9 & \underline{87.2} & 71.6 & \textbf{79.2} \\
  \rowcolor{mixsdGreen}\cellcolor{white}& \methodname{} ($\lambda{=}0.5$) & 56.6 & \textbf{79.0} & \underline{52.1} & \textbf{93.4} & \underline{92.4} & 84.8 & \textbf{73.0} & \underline{79.1} \\
\midrule
\multirow{7}{*}{Qwen3-8B}
  & Base                            &  2.9 & 65.4 & 26.0 & 83.4 & 91.8 & 86.6 & 73.1 & 72.2 \\
  & SFT                             & \underline{86.9} &  2.4 &  5.4 & 41.6 & 63.2 & \underline{84.1} & \textbf{73.1} & 53.5 \\
  & OPSD                            & 80.0 & 25.2 & 15.6 & 76.6 & \underline{90.8} & \textbf{86.6} & 56.8 & \underline{65.3} \\
  & OPSD-NLL ($n{=}1$, $T{=}0$)     & \underline{86.9} & 12.4 & 12.5 & 74.8 & 88.9 & 81.1 & 53.8 & 62.2 \\
  & \methodname{} ($\lambda{=}0$)   & 78.3 &  6.4 & 14.4 & 74.2 & 88.6 & \underline{84.1} & 43.3 & 60.9 \\
  \rowcolor{mixsdBlue}\cellcolor{white}& \methodname{} ($\lambda{=}0.3$) & \textbf{89.1} & \underline{28.6} & \textbf{21.2} & \underline{78.2} & 90.3 & 82.3 & \underline{60.7} & \textbf{66.6} \\
  \rowcolor{mixsdGreen}\cellcolor{white}& \methodname{} ($\lambda{=}0.5$) & 52.0 & \textbf{31.6} & \underline{20.4} & \textbf{79.2} & \textbf{91.3} & \underline{84.1} & 57.8 & \textbf{66.6} \\
\bottomrule
\end{tabular}
}
\end{table}

We evaluate \methodname{} across four training corpora and three model scales, comparing against SFT and OPSD. Across all settings, we observe a clear trade-off between memorization of injected knowledge and preservation of pre-existing capabilities.

\paragraph{SFT achieves strong memorization but causes severe forgetting.}
SFT achieves near-perfect performance on training objectives but substantially degrades performance on held-out capability benchmarks. On \datasetasmall{} (Table~\ref{tab:knowledge-eval-small}), SFT attains high training performance while reducing the average held-out capability score by $30$-$40\%$. On \datasetb{} (Table~\ref{tab:math-op-eval}), SFT performs well on in-domain test accuracy but nearly collapses generalization to unseen operations (\datasetbnewops{}). Table~\ref{tab:knowledge-eval-large} and Table~\ref{tab:simpleqa-eval} in Appendix~\ref{appn:additional_results} show similar trends on \datasetalarge{} and SimpleQA, respectively. Held-out capability benchmarks show the same degradation trend across all model scales. These results indicate that SFT memorizes new knowledge at the cost of disrupting existing capabilities.

OPSD often preserves more capability than SFT, but its performance is inconsistent across datasets and model scales, as also observed in prior work~\citep{kim2026does}. For example, on \datasetasmall{} with Qwen3-1.7B, OPSD achieves an average held-out capability score of only $5.1$, below SFT’s $14.3$. Moreover, our default OPSD setting samples eight rollouts per prompt ($n{=}8$), making it substantially more computationally expensive due to repeated on-policy generation and more data-hungry to train effectively.

\paragraph{\methodname{} improves the injection-retention trade-off.}

\begin{figure}[t]
\centering
\includegraphics[width=\linewidth]{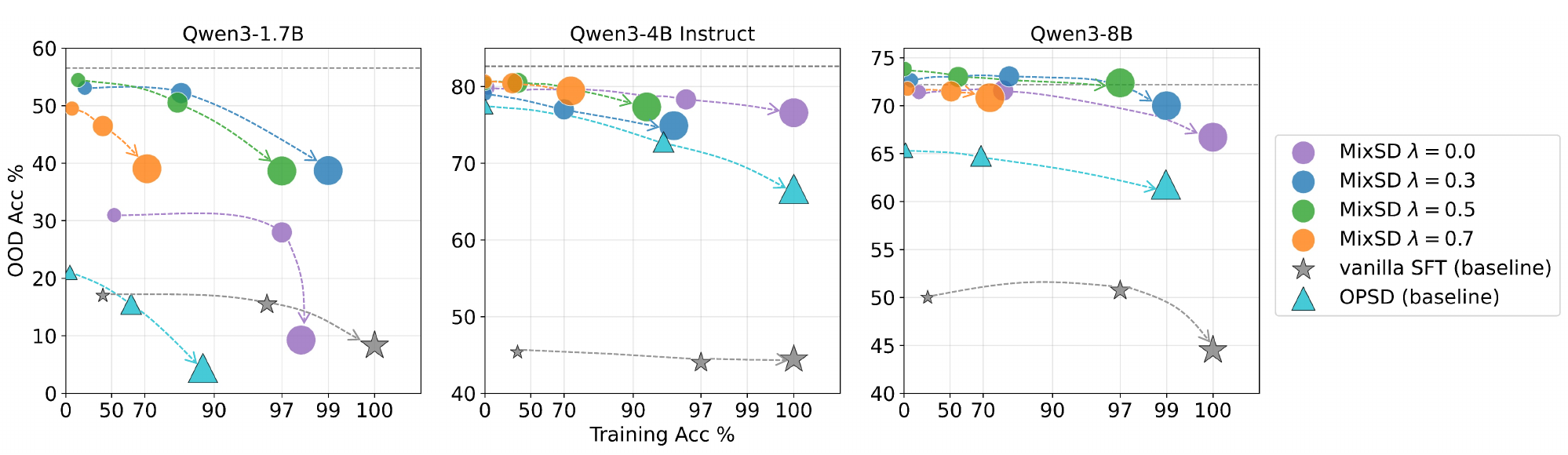}
\caption{Trade-off between training accuracy on \datasetasmall{} and average general-domain OOD test accuracy across AIME2024, MATH500, GSM8K, HumanEval, and MMLU. Each point corresponds to a checkpoint at a different training step, with larger markers indicating later stages of training. The horizontal dashed lines denote the average OOD accuracy of the untrained base model. We observe a consistent trade-off between training and OOD performance across models and methods, while \methodname{} exhibits reduced forgetting compared to the baselines.}
\label{fig:pareto-d5-e10}
\end{figure}

Across all datasets and model scales, \methodname{} maintains strong training performance while preserving substantially more of the model’s existing capabilities. Figure~\ref{fig:pareto-d5-e10} illustrates this effect on \datasetasmall{}, where \methodname{} traces a substantially better Pareto frontier between memorization of injected knowledge and held-out capability retention. On \datasetasmall{}, \methodname{} ($\lambda{=}0.3$ or $0.5$) achieves high training accuracy while significantly improving the average held-out capability score. Similar trends hold on \datasetalarge{} (Table~\ref{tab:knowledge-eval-large}) and SimpleQA (Table~\ref{tab:simpleqa-eval}), where \methodname{} consistently outperforms both SFT and OPSD on held-out capability benchmarks.

We also observe a clear scaling effect: although \methodname{} consistently improves over SFT at all model sizes, larger models exhibit substantially less forgetting than smaller ones. We hypothesize that effectively learning from mixed supervision requires sufficient existing capability to integrate tokens sampled from different conditional distributions without destabilizing generation behavior.

\paragraph{The mixing rate $\lambda$ controls the injection-retention trade-off.}

The mixing rate $\lambda$ provides a simple mechanism for balancing memorization of injected knowledge against retention of existing capabilities. Smaller values emphasize expert-conditioned supervision and favor memorization, while larger values introduce more naive tokens that anchor the model to its prior. This trade-off is consistent across datasets. As shown in \cref{tab:knowledge-eval-small}, \cref{tab:math-op-eval} and Appendix~\ref{appn:additional_results}, increasing $\lambda$ from $0$ to $0.5$ substantially improves held-out capability retention with only a modest reduction in training accuracy, while $\lambda{=}0.7$ begins to noticeably degrade memorization.

\section{Discussion}
\label{sec:discussion}

We analyze why and how broadly \methodname{}'s memorization and retention gains hold along five axes: the parameter space mechanism behind forgetting (\cref{sec:discussion_fisher}), how mixing reshapes the token level training signal (\cref{sec:discussion_reshape}), the qualitative error modes induced by each method (\cref{sec:discussion_errors}), cross family generality on a Llama backbone (\cref{sec:discussion_llama}), and transfer to the knowledge editing setting (\cref{sec:discussion_editing}).

\subsection{Fine-tuning direction matters more than update magnitude}
\label{sec:discussion_fisher}

We study whether catastrophic forgetting is better explained by the direction of parameter updates rather than their magnitude. To characterize sensitive directions in the base model, we use the Fisher information matrix,
\begin{equation}
F(\theta) =
\mathbb{E}_{\mathbf{x}, \mathbf{y}}
\left[
\nabla_{\theta} \log p(\mathbf{y} \mid \mathbf{x}, \theta)
\nabla_{\theta} \log p(\mathbf{y} \mid \mathbf{x}, \theta)^{\top}
\right],
\end{equation}
which captures the local curvature of the model’s log-likelihood landscape. We approximate $F$ using the diagonal empirical Fisher~\citep{kirkpatrick2017overcoming}:
\begin{equation}
\hat{F}_i =
\frac{1}{N} \sum_{n=1}^{N}
\left(
\nabla_{\theta_i} \log p(\mathbf{y}_n \mid \mathbf{x}_n, \theta)
\right)^2,
\end{equation}
where $\theta_i$ denotes the $i$-th model parameter. Large $\hat{F}_i$ values correspond to parameters whose perturbation strongly affects the model’s likelihood, indicating directions that are particularly important to the base model. We estimate $\hat F$ for each base model using a random subset of $N{=}979$ examples drawn from the five general-domain benchmarks.

To measure how strongly fine-tuning updates align with these sensitive directions, we define the Fisher alignment ratio
\begin{equation}
R \;=\;
\frac{\Delta\theta^{\top} \hat{F}\, \Delta\theta\,/\,\|\Delta\theta\|^{2}}
     {\operatorname{tr}(\hat{F})\,/\,d},
\end{equation}
which compares the average curvature along the update direction to the expected curvature under an isotropic direction. Values $R > 1$ indicate that updates concentrate in sensitive, high-curvature directions that are more likely to induce forgetting, while $R < 1$ indicates that they preferentially avoid them.

We compare this directional measure with the raw weight displacement $\|\Delta\theta\|^2$, which captures only the magnitude of parameter change. Additional details and results for both metrics across methods and model sizes are provided in \cref{app:fisher-info}. We find that displacement magnitude alone does not reliably predict forgetting ($r{=}+0.34$, $+0.02$, and $+0.10$ for 1.7B, 4B, and 8B models, respectively; $n \approx 40$ checkpoints per size): SFT and \methodname{} often exhibit similar parameter displacement while yielding substantially different levels of capability degradation. In contrast, the alignment ratio $R$ is strongly correlated with forgetting within each model size ($r{=}+0.56$, $+0.82$, and $+0.57$ for 1.7B, 4B, and 8B models, respectively). This suggests that update direction is a much stronger predictor of forgetting than update magnitude.

\subsection{How does mixing reshape the token-level training signal?}
\label{sec:discussion_reshape}

\begin{figure}[t]
\centering
\includegraphics[width=\linewidth]{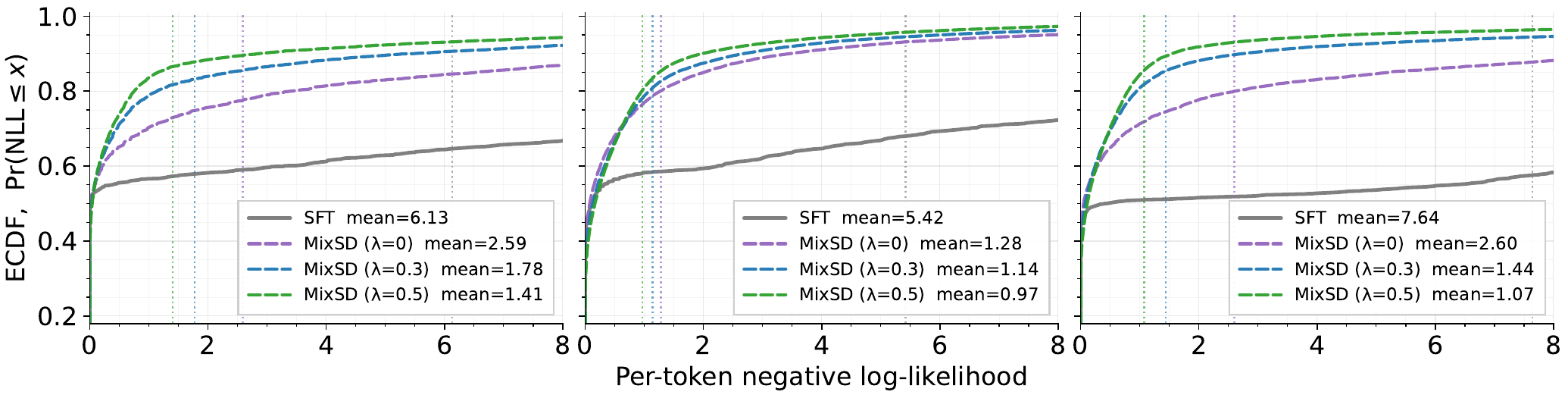}
\caption{Empirical CDFs of per-token negative log-likelihood (NLL) under the base model, evaluated on training targets from standard SFT (raw ground-truth) and from \methodname{} at mixing rates $\lambda \in \{0, 0.3, 0.5\}$. \emph{Left}: Qwen3-1.7B. \emph{Middle}: Qwen3-4B-Instruct. \emph{Right}: Qwen3-8B. Vertical dotted lines denote the mean NLL for each method. Across all model scales, \methodname{} produces training targets with substantially lower NLL than raw ground-truth targets, and increasing $\lambda$ progressively shifts the targets further toward the model’s naive self-distribution and reference mode.}
\label{fig:nll-d5-e10}
\end{figure}

Figure~\ref{fig:nll-d5-e10} shows the empirical CDF of per-token negative log-likelihood (NLL) under the base model for SFT and \methodname{} targets. Across all model scales, \methodname{} shifts the target distribution toward substantially lower NLL, with larger $\lambda$ producing stronger shifts as more tokens are sampled from the model’s own rollout distribution. To quantify this effect, we consider high-NLL tokens whose base-model NLL exceeds 8. On the knowledge datasets, such tokens constitute 27--42\% of SFT targets, but only 4--8\% for \methodname{} at $\lambda{=}0.3$, falling further to 2--3\% at $\lambda{=}0.7$. Similar trends hold across datasets, model scales, and NLL thresholds (Appendix~\ref{app:nll}). Importantly, \methodname{} largely preserves the core memorization signal: 81--98\% of the unique high-NLL token types appearing in SFT targets also appear in the corresponding \methodname{} targets (Table~\ref{tab:overlap-recall}). Thus, \methodname{} reduces the overall amount of low-probability supervision while retaining most tokens that encode genuinely new information.

\subsection{What kinds of errors does each method make?}
\label{sec:discussion_errors}

Beyond aggregate accuracy, SFT and \methodname{} exhibit qualitatively different failure modes on held-out benchmarks. We classify incorrect responses into four mutually exclusive categories: \texttt{format} (no parseable answer), \texttt{leakage} (generation of fictional \dataseta{} entities unrelated to the prompt), \texttt{collapse} (short template-style answers without reasoning), and \texttt{genuine} (coherent but incorrect reasoning attempts). On Qwen3-1.7B fine-tuned on \datasetalarge{}, $50.7\%$ of all $14{,}042$ MMLU errors under SFT fall into the \texttt{leakage} category, and another $48.0\%$ into \texttt{collapse}; only $0.4\%$ correspond to genuine reasoning errors. Under \methodname{}, these pathological modes nearly disappear: \texttt{leakage} and \texttt{collapse} account for at most $3.4\%$ and $0.6\%$ of errors respectively, while the majority ($\geq 71\%$) are \texttt{genuine}, closely matching the base model’s error distribution. These results suggest that SFT does not merely reduce held-out accuracy, but qualitatively changes the model’s generation behavior, causing unrelated prompts to trigger memorized artifacts from the fine-tuning corpus. In contrast, \methodname{} largely preserves the base model’s original error profile. Full per-method and per-benchmark breakdowns for \datasetasmall{} and \datasetb{} are provided in Appendix~\ref{appn:error-analysis}.

\subsection{Does the effect transfer across model families?}
\label{sec:discussion_llama}

We additionally evaluate Llama-3.2-1B-Instruct on \datasetasmall{} (Appendix~\ref{appn:llama-1b}). The same qualitative trends hold: SFT achieves near-perfect training accuracy but severely degrades held-out capabilities, whereas \methodname{} at moderate mixing rates ($\lambda{=}0.3$-$0.5$) preserves substantially more of the model’s original performance. In particular, \methodname{} with $\lambda{=}0.5$ retains 78\% of the base model’s average held-out score while achieving 98\% training accuracy, compared to only 21\% retention for SFT. These results suggest that the benefits of mixing teacher-conditioned and self-generated targets are not specific to the Qwen family, but reflect a more general property of the training signal.

\subsection{Does \methodname{} transfer to knowledge editing?}
\label{sec:discussion_editing}

Our main experiments inject entirely novel facts into the model. To test whether \methodname{} also applies when existing knowledge must be \emph{revised}, we fine-tune on a random subset of the MQuAKE dataset~\citep{zhong2023mquake}, which requires overwriting facts already stored in the model’s parameters. We evaluate all three Qwen3 scales (1.7B, 4B, 8B) using the same training protocol; full results are provided in Appendix~\ref{appn:mquake}. The qualitative trends remain consistent with the knowledge-injection setting. SFT rapidly memorizes the edited facts but induces severe forgetting, reducing the held-out capability average to 7.8--39.4\%. In contrast, \methodname{} at $\lambda{=}0.3$ achieves comparable editing accuracy while retaining substantially more general capability, preserving above 90\% of held-out performance on the 4B and 8B models. We additionally compare against MEMIT~\citep{meng2023massediting}, a locate-and-edit method specifically designed for knowledge editing. While MEMIT is not applicable in our knowledge-injection setting, it is naturally suited to MQuAKE. In practice, however, it exhibits the opposite trade-off: held-out capabilities remain nearly unchanged, but editing accuracy is substantially lower (53--70\% across scales) than either SFT or \methodname{}. This limitation likely arises because knowledge editing modifies many related facts simultaneously: entities shared across multiple knowledge triples cause the rank-one updates to interfere destructively (Table~\ref{tab:mquake-full}).

\section{Conclusion}
\label{sec:conclusion}

We presented \methodname{}, a simple external-teacher-free method for forgetting-aware knowledge injection in large language models. Our results suggest that catastrophic forgetting arises in part from a mismatch between externally authored supervision targets and the model’s native autoregressive distribution. By constructing distribution-aligned supervision through mixed self-generated targets, \methodname{} consistently achieves a substantially improved memorization-retention trade-off across multiple datasets, model scales, and knowledge injection settings. We hope this work motivates future research on distribution-aware supervision and continual adaptation for large language models.

\section{Limitations}
\label{app:limitation}

\methodname{} introduces an additional hyperparameter $\lambda$; while $\lambda=0.3$ consistently performs well in our experiments, the optimal value may differ for other tasks. Our largest backbone is 8B parameters, and how the method scales to substantially larger models (70B and beyond) remains an open question. Finally, generating mixed rollouts incurs a one-time preprocessing cost over standard SFT, though this is substantially smaller than the per-step cost of on-policy self-distillation.

\section*{Acknowledgment}
We thank Weihua Du at the Carnegie Mellon University for initial discussions and insights on this topic. This material is based in part upon work supported by the German Federal Ministry of Education and Research (BMBF): Tübingen AI Center, FKZ: 01IS18039B; by the Machine Learning Cluster of Excellence, EXC number 2064/1 – Project number 390727645; and by Schmidt Sciences SAFE-AI Grant.

\bibliographystyle{plainnat}
\bibliography{custom}


\newpage
\appendix

\section{Experimental Results on \datasetalarge{} and SimpleQA}
\label{appn:additional_results}

We present the experimental results on \datasetalarge{} and SimpleQA in \cref{tab:knowledge-eval-large} and \cref{tab:simpleqa-eval}, respectively.

\begin{table}[h]
\centering
\caption{Evaluation on \datasetalarge{} across the three model backbones. Methods and column semantics follow Table~\ref{tab:knowledge-eval-small}. All values are in percent. \textbf{Bold} and \underline{underline} mark the best and second-best method per column (excluding Base).}
\label{tab:knowledge-eval-large}
\setlength{\tabcolsep}{4pt}
\resizebox{\textwidth}{!}{
\begin{tabular}{ll c c cccccc}
\toprule
\multirow{2}{*}{\textbf{Model}} & \multirow{2}{*}{\textbf{Method}} & \multicolumn{2}{c}{\textbf{In-domain test}} & \multicolumn{6}{c}{\textbf{Held-out capability test}} \\
 \cmidrule(lr){3-4} \cmidrule(lr){5-10}
 &  & \textbf{Train} & \textbf{\datasetaretrieval{}} & \textbf{AIME-2024} & \textbf{MATH-500} & \textbf{GSM8K} & \textbf{HumanEval} & \textbf{MMLU} & \textbf{Avg} \\
\midrule
\multirow{6}{*}{Qwen3-1.7B}
  & Base                            &  0.0 & 97.3 & 11.0 & 72.4 & 80.4 & 60.4 & 58.5 & 56.5 \\
  & SFT                             & \textbf{99.6} &  0.0 &  0.0 &  1.8 &  0.4 &  0.0 &  0.4 &  0.5 \\
  & OPSD                            & 91.9 &  1.3 &  0.0 &  1.8 &  0.8 &  0.6 &  4.4 &  1.5 \\
  & \methodname{} ($\lambda{=}0$)   & 95.2 & \underline{12.8} &  0.0 &  5.2 &  3.0 &  0.0 &  9.4 &  3.5 \\
  \rowcolor{mixsdBlue}\cellcolor{white}& \methodname{} ($\lambda{=}0.3$) & \underline{96.5} & \textbf{14.3} & \underline{0.2} & \underline{13.6} & \underline{13.2} & \underline{4.9} & \textbf{24.6} & \underline{11.3} \\
  \rowcolor{mixsdGreen}\cellcolor{white}& \methodname{} ($\lambda{=}0.5$) & 95.6 &  4.8 & \textbf{0.6} & \textbf{26.8} & \textbf{32.1} & \textbf{20.7} & \underline{23.5} & \textbf{20.7} \\
\midrule
\multirow{5}{*}{Qwen3-4B-It}
  & Base                            &  0.0 & 97.9 & 62.9 & 94.2 & 92.6 & 86.0 & 77.6 & 82.6 \\
  & SFT                             & \underline{97.1} & 36.8 &  5.2 & 30.0 & 22.7 & 81.1 & 56.9 & 39.2 \\
  & \methodname{} ($\lambda{=}0$)   & \textbf{98.1} & \textbf{98.7} & 23.5 & 80.8 & \underline{88.0} & \underline{84.8} & \textbf{65.5} & 68.5 \\
  \rowcolor{mixsdBlue}\cellcolor{white}& \methodname{} ($\lambda{=}0.3$) & 94.0 & 96.4 & \underline{41.5} & \underline{84.8} & 87.2 & 83.5 & 53.7 & \underline{70.1} \\
  \rowcolor{mixsdGreen}\cellcolor{white}& \methodname{} ($\lambda{=}0.5$) & 84.6 & \underline{97.9} & \textbf{44.8} & \textbf{93.0} & \textbf{91.7} & \textbf{87.2} & \underline{61.0} & \textbf{75.5} \\
\midrule
\multirow{5}{*}{Qwen3-8B}
  & Base                            &  0.0 & 99.6 & 26.0 & 83.4 & 91.8 & 86.6 & 73.1 & 72.2 \\
  & SFT                             & \underline{98.3} & 71.4 &  7.7 & 34.4 & 25.4 & \textbf{83.5} & \underline{69.9} & 44.2 \\
  & \methodname{} ($\lambda{=}0$)   & \textbf{98.5} & \textbf{94.9} & 10.6 & 69.8 & \underline{89.2} & 61.0 & 67.3 & 59.6 \\
  \rowcolor{mixsdBlue}\cellcolor{white}& \methodname{} ($\lambda{=}0.3$) & 95.8 & \underline{84.2} & \underline{22.5} & \underline{77.6} & \textbf{91.4} & 79.3 & 67.4 & \underline{67.6} \\
  \rowcolor{mixsdGreen}\cellcolor{white}& \methodname{} ($\lambda{=}0.5$) & 93.1 & 65.3 & \textbf{24.2} & \textbf{82.0} & \textbf{91.4} & \underline{81.1} & \textbf{71.0} & \textbf{69.9} \\
\bottomrule
\end{tabular}
}
\end{table}

\begin{table}[h]
\centering
\caption{Evaluation on SimpleQA across the three model backbones. The Train column reports closed-book accuracy on the 4{,}326-question SimpleQA training split. The Held-out capability test columns report five unrelated benchmarks that probe forgetting; Avg is their unweighted mean. All values are in percent. \textbf{Bold} and \underline{underline} mark the best and second-best method per column (excluding Base).}
\label{tab:simpleqa-eval}
\setlength{\tabcolsep}{4pt}
\resizebox{\textwidth}{!}{
\begin{tabular}{ll c cccccc}
\toprule
\multirow{2}{*}{\textbf{Model}} & \multirow{2}{*}{\textbf{Method}} & \textbf{Train} & \multicolumn{6}{c}{\textbf{Held-out capability test}} \\
\cmidrule(lr){3-3} \cmidrule(lr){4-9}
 &  & \textbf{SimpleQA} & \textbf{AIME-2024} & \textbf{MATH-500} & \textbf{GSM8K} & \textbf{HumanEval} & \textbf{MMLU} & \textbf{Avg} \\
\midrule
\multirow{7}{*}{Qwen3-1.7B}
  & Base                            &  0.5 & 11.0 & 72.4 & 80.4 & 60.4 & 58.5 & 56.5 \\
  & SFT                             & \textbf{8.5} &  0.0 &  6.8 &  3.4 &  3.0 & \underline{30.2} &  8.7 \\
  & OPSD                            &  2.9 &  0.4 & 19.0 & 24.3 & 31.1 &  4.7 & 15.9 \\
  & OPSD ($n{=}1$, $T{=}0$)         &  3.9 &  0.0 &  6.4 &  7.2 &  8.5 &  6.9 &  5.8 \\
  & OPSD-NLL ($n{=}1$, $T{=}0$)     &  3.4 & \textbf{1.3} & 18.0 & 18.3 & 18.9 & 10.8 & 13.4 \\
  \rowcolor{mixsdBlue}\cellcolor{white}& \methodname{} ($\lambda{=}0.3$) & \underline{7.1} &  0.6 & \underline{29.8} & \underline{37.1} & \underline{33.5} & 20.9 & \underline{24.4} \\
  \rowcolor{mixsdGreen}\cellcolor{white}& \methodname{} ($\lambda{=}0.5$) &  6.3 & \underline{1.0} & \textbf{31.2} & \textbf{45.2} & \textbf{36.0} & \textbf{31.4} & \textbf{29.0} \\
\midrule
\multirow{7}{*}{Qwen3-4B-It}
  & Base                            &  3.4 & 62.9 & 94.2 & 92.6 & 86.0 & 77.6 & 82.6 \\
  & SFT                             & \textbf{18.8} &  0.2 & 19.2 & 15.0 & 50.0 & 48.8 & 26.6 \\
  & OPSD                            &  7.5 & 33.8 & 73.2 & 77.0 & \underline{88.4} & \textbf{67.1} & 67.9 \\
  & OPSD ($n{=}1$, $T{=}0$)         &  8.5 & 46.0 & 82.8 & 83.3 & 81.7 & \underline{62.2} & 71.2 \\
  & OPSD-NLL ($n{=}1$, $T{=}0$)     &  7.8 & 32.5 & 62.2 & 55.4 & 79.3 & 49.6 & 55.8 \\
  \rowcolor{mixsdBlue}\cellcolor{white}& \methodname{} ($\lambda{=}0.3$) & \underline{10.5} & \underline{55.4} & \underline{92.6} & \underline{90.1} & \textbf{89.0} & 53.9 & \textbf{76.2} \\
  \rowcolor{mixsdGreen}\cellcolor{white}& \methodname{} ($\lambda{=}0.5$) & 10.4 & \textbf{61.9} & \textbf{93.2} & \textbf{90.5} & 84.8 & 49.3 & \underline{75.9} \\
\midrule
\multirow{7}{*}{Qwen3-8B}
  & Base                            &  2.5 & 26.0 & 83.4 & 91.8 & 86.6 & 73.1 & 72.2 \\
  & SFT                             & \textbf{16.8} &  0.2 & 18.8 & 14.1 & 67.1 & 52.0 & 30.4 \\
  & OPSD                            &  8.3 & 14.2 & 75.0 & 80.8 & 79.3 & 59.4 & 61.7 \\
  & OPSD ($n{=}1$, $T{=}0$)         &  6.8 & 19.0 & \underline{80.2} & 89.3 & \underline{85.4} & 62.2 & 67.2 \\
  & OPSD-NLL ($n{=}1$, $T{=}0$)     &  6.0 & 20.6 & 79.0 & \underline{89.4} & \textbf{86.6} & \textbf{70.2} & \underline{69.2} \\
  \rowcolor{mixsdBlue}\cellcolor{white}& \methodname{} ($\lambda{=}0.3$) &  9.0 & \underline{21.9} & 79.4 & 87.0 & 81.1 & 65.0 & 66.9 \\
  \rowcolor{mixsdGreen}\cellcolor{white}& \methodname{} ($\lambda{=}0.5$) & \underline{9.3} & \textbf{27.1} & \textbf{80.8} & \textbf{92.3} & 82.9 & \underline{66.8} & \textbf{70.0} \\
\bottomrule
\end{tabular}
}
\end{table}

\section{Additional Related Work: Evaluation Datasets for Knowledge Injection}
Evaluation for knowledge injection must distinguish genuine acquisition of new information from memorization, annotation artifacts, or noisy supervision. Real-world factuality benchmarks such as SimpleQA provide short, information-seeking questions with objectively verifiable answers~\citep{wei2024measuring}, making them useful for testing factual recall after fine-tuning, but they do not provide controlled target knowledge to inject. Knowledge editing benchmarks address this limitation by constructing explicit factual updates: ZsRE and CounterFact test whether models can revise individual facts while preserving locality~\citep{levy2017zero,decao2021editing,meng2022locating}, while MQuAKE and RippleEdits further evaluate whether such updates propagate to logically related or multi-hop questions~\citep{zhong2023mquake,cohen2024evaluating}. However, these datasets often rely on real-world facts, knowledge bases, or generated counterfactual variants, so evaluation can still be confounded by pretraining contamination, ambiguous entities, outdated facts, and construction noise. Broader suites such as DUnE and KnowEdit improve coverage across domains and evaluation dimensions, but they do not fully remove the difficulty of validating factual supervision at scale~\citep{akyurek2023dune,zhang2024comprehensive}. Recent synthetic benchmarks such as PhantomWiki instead generate fictional worlds on demand, allowing reasoning and retrieval to be evaluated against known ground truth rather than memorized or hallucinated facts~\citep{gong2025phantomwiki}. Motivated by this principle, our datasets provide a clean and fully verifiable setting for studying knowledge injection and forgetting without relying on facts that may already exist in the pretrained model.

\section{Broader Impacts}
\label{appn:impacts}

\methodname{} is a general-purpose fine-tuning method for injecting knowledge into language models while mitigating catastrophic forgetting. By improving the memorization-retention trade-off, it can potentially lower the data and compute cost of adapting LLMs to specialized domains such as medicine, law, and low-resource languages, making domain adaptation more accessible, and may potentially help preserve safety-relevant behaviors (e.g., instruction following, refusal calibration) that are often degraded by standard fine-tuning. As with any fine-tuning method, it could in principle be used to inject misleading or harmful content into a base model; however, because \methodname{} requires only a base model and a target corpus and introduces no new misuse vectors beyond those of standard SFT, we consider its marginal misuse risk to be low.

\section{Additional Experiment Setup Details}
\label{appn:details}

We use \url{https://github.com/NVIDIA-NeMo/RL} for all training experiments. We set the rollout number $n$ to 1 for SFT and \methodname{}, and to 8 for OPSD unless otherwise specified. For OPSD, we use the forward KL objective and keep the teacher model fixed as the base model without parameter updates. We also evaluate additional OPSD variants by replacing the KL loss with an NLL loss and by varying the number of rollouts per query $n$ and the rollout-generation temperature $T$. We set the top-$k$ logits parameter to $k=64$.

For \methodname{}, we allow up to 10 retries if the generated output is identified as incorrect by our rule-based verifier. If all retries fail, the example is discarded. In practice, we find that this retry budget is sufficient for \methodname{} to generate correct outputs on both \dataseta{} and \datasetb{}. For SimpleQA, examples that remain incorrect after all retries are similarly discarded.

We tune the learning rate over $\{1\times10^{-6}, 5\times10^{-6}, 1\times10^{-5}, 3\times10^{-5}, 5\times10^{-5}, 1\times10^{-4}\}$ and select the best learning rate based on validation performance for each training run. The global batch size is 16, and training is performed in bfloat16 precision. We use a warmup scheduler with 20 warmup steps, followed by a constant learning rate. We set the number of prompts per step to 20. Checkpoints are saved every 20 steps for \dataseta{} and SimpleQA, and every 10 steps for \datasetb{}.

We select the earliest checkpoint after convergence, defined as the first checkpoint that achieves the best training performance observed so far while maintaining stable validation performance, i.e., without a validation performance drop greater than 5\% over the following two consecutive checkpoints.

For \dataseta{}, we train for 20 epochs. For \datasetb{}, we train for 100 steps. For SimpleQA, we train for one epoch over the full dataset.

For rollout generation, we use VLLM V1. We set the maximum number of new tokens to 8192 and the total sequence length to 10{,}000. Unless otherwise noted, rollouts are generated with temperature $T=0$, so token-proposal randomness arises solely from Bernoulli mixing. All experiments are conducted using 4$\times$H100 GPUs, with a total training cost of approximately 2000 GPU hours.

\section{Additional Experiment Results}

\subsection{Fisher Information: Estimation and Diagnostics}
\label{app:fisher-info}

This section describes the Fisher approximation, estimation procedure, and diagnostic quantities used in Section~\ref{sec:discussion}.

\paragraph{Estimation procedure}

We estimate $\hat{F}$ at the base model $\theta^\star$ using a reference corpus $\mathcal{D}_{\mathrm{ref}}$ of $N{=}979$ examples randomly drawn from five general-domain benchmarks (AIME-2024, MATH-500, GSM8K, HumanEval, and MMLU).

For each example $(\mathbf{x}_n, \mathbf{y}_n)$:
\begin{enumerate}
\item Compute the sum-reduced log-likelihood over target tokens (prompt tokens masked).
\item Backpropagate to obtain the per-sample gradient $\mathbf{g}_n$.
\item Accumulate $\mathbf{g}_n^2$ into the Fisher estimate.
\end{enumerate}

We use batch size 1 to obtain per-sample gradients, and cast gradients to fp32 before squaring to avoid underflow. After processing all samples, we average over $N$.

\paragraph{Diagnostic quantities}

Given a fine-tuned checkpoint $\theta_B$ with displacement $\Delta\theta = \theta_B - \theta^\star$, we compute:

Raw parameter displacement:
\begin{equation}
\|\Delta\theta\|^2 = \sum_i (\Delta\theta_i)^2.
\end{equation}

Fisher-weighted displacement:
\begin{equation}
Q_F = \Delta\theta^\top \hat{F} \Delta\theta = \sum_i \hat{F}_i (\Delta\theta_i)^2.
\end{equation}

$Q_F$ measures how strongly the update moves along sensitive directions of the base model.

Fisher alignment ratio:
\begin{equation}
R =
\frac{Q_F / \|\Delta\theta\|^2}
{\operatorname{tr}(\hat{F}) / d}.
\end{equation}

$R$ compares curvature along the update direction to the expected curvature under an isotropic direction. Values $R{>}1$ indicate concentration in high-curvature directions, while $R{<}1$ indicates avoidance.

\paragraph{Parameter displacement across methods and model sizes}
\label{app:fisher-displacement}
\label{app:fisher-correlation}

Table~\ref{tab:app-displacement} reports the raw parameter displacement $\|\Delta\theta\|^2$ and Fisher alignment ratio $R$ for each method, model size, and task cohort.

\begin{table}[h]
\centering
\caption{Raw parameter displacement $\|\Delta\theta\|^2$, Fisher alignment ratio $R$, and mean capability drop (Drop: average accuracy decrease across AIME-2024, MATH-500, GSM8K, HumanEval, and MMLU relative to the base model) at converged checkpoints for each method, model size, and task cohort. Methods with similar displacement can produce dramatically different forgetting (Drop column), confirming that magnitude alone does not predict capability loss.}
\label{tab:app-displacement}
\setlength{\tabcolsep}{4pt}
\resizebox{\textwidth}{!}{
\begin{tabular}{ll rrr rrr rrr}
\toprule
& & \multicolumn{3}{c}{\textbf{\datasetasmall{}}} & \multicolumn{3}{c}{\textbf{\datasetb{}}} & \multicolumn{3}{c}{\textbf{SimpleQA}} \\
\cmidrule(lr){3-5} \cmidrule(lr){6-8} \cmidrule(lr){9-11}
\textbf{Model} & \textbf{Method} & $\|\Delta\theta\|^2$ & $R$ & Drop & $\|\Delta\theta\|^2$ & $R$ & Drop & $\|\Delta\theta\|^2$ & $R$ & Drop \\
\midrule
\multirow{5}{*}{Qwen3-1.7B}
  & SFT                             &   85.0 & 0.90 & 42.3 &  58.0 & 1.10 & 53.3 &  469.9 & 0.77 & 47.9 \\
  & OPSD-fwdKL                      & 1752.4 & 0.49 & 51.5 &  62.5 & 0.58 & 13.8 & 2332.1 & 0.43 & 40.6 \\
  \rowcolor{mixsdBlue}\cellcolor{white}& \methodname{} ($\lambda{=}0.3$) &  215.2 & 0.67 & 16.7 &  69.4 & 0.78 & 14.3 &  430.3 & 0.65 & 32.1 \\
  \rowcolor{mixsdGreen}\cellcolor{white}& \methodname{} ($\lambda{=}0.5$) &  221.1 & 0.65 & 16.3 &  65.0 & 0.75 & 10.9 &  424.7 & 0.67 & 27.6 \\
\midrule
\multirow{5}{*}{Qwen3-4B-It}
  & SFT                             &  28.2 & 0.99 & 39.4 &  16.6 & 1.14 & 66.0 &   96.7 & 1.36 & 56.0 \\
  & OPSD-fwdKL                      & 346.5 & 0.81 & 22.0 &  23.6 & 0.74 &  5.0 &  485.5 & 0.64 & 14.8 \\
  \rowcolor{mixsdBlue}\cellcolor{white}& \methodname{} ($\lambda{=}0.3$) &  63.5 & 0.70 &  8.4 &  20.9 & 0.86 &  3.4 &   82.2 & 0.83 &  6.5 \\
  \rowcolor{mixsdGreen}\cellcolor{white}& \methodname{} ($\lambda{=}0.5$) &  68.5 & 0.65 &  5.1 &  18.7 & 0.81 &  3.5 &   81.8 & 0.82 &  6.7 \\
\midrule
\multirow{5}{*}{Qwen3-8B}
  & SFT                             &  61.9 & 1.16 & 26.0 &  53.9 & 1.25 & 18.7 &  187.6 & 2.14 & 41.8 \\
  & OPSD-fwdKL                      & 577.7 & 0.91 & 10.4 &  44.3 & 0.99 &  6.9 & 1109.6 & 0.96 & 10.5 \\
  \rowcolor{mixsdBlue}\cellcolor{white}& \methodname{} ($\lambda{=}0.3$) &  88.6 & 1.00 &  2.2 &  42.6 & 1.39 &  5.6 &  159.3 & 1.35 &  5.3 \\
  \rowcolor{mixsdGreen}\cellcolor{white}& \methodname{} ($\lambda{=}0.5$) &  86.5 & 0.92 & $-$0.2 & 40.5 & 1.33 &  5.6 &  161.5 & 1.48 &  2.2 \\
\bottomrule
\end{tabular}
}
\end{table}

Table~\ref{tab:app-corr-pooled} reports the Pearson correlation between each metric ($\|\Delta\theta\|^2$ and $R$) and mean forgetting (average drop across five held-out benchmarks), pooling all checkpoints across the three task cohorts.

\begin{table}[h]
\centering
\caption{Pearson $r$ between each metric and forgetting.}
\label{tab:app-corr-pooled}
\begin{tabular}{l c cc}
\toprule
\textbf{Model} & $n$ & $r(\|\Delta\theta\|^2, \text{Drop})$ & $r(R, \text{Drop})$ \\
\midrule
Qwen3-1.7B & 42 & $+$0.34 & $+$0.56 \\
Qwen3-4B-It   & 45 & $+$0.02 & $+$0.82 \\
Qwen3-8B   & 40 & $+$0.10 & $+$0.57 \\
\bottomrule
\end{tabular}
\end{table}

\subsection{High-NLL Token Counts}
\label{app:nll}

Since NLL $= -\log p$, a token with NLL $= \tau$ receives predicted probability $p = e^{-\tau}$ from the base model.
At $\tau = 5$ this corresponds to $p \approx 0.7\%$---the model is highly uncertain but retains weak signal---while at $\tau = 8$ the probability drops to $p \approx 0.034\%$, indicating near-complete failure of prediction (equivalent to uniform guessing among ${\sim}3{,}000$ alternatives).
We report counts at both thresholds: Table~\ref{tab:high-nll-5} captures tokens where the base model is very uncertain, and Table~\ref{tab:high-nll-8} captures tokens where it has essentially no predictive signal.
The relative ordering across training configurations is consistent at both thresholds, confirming that the findings are not sensitive to the specific choice of $\tau$.

\begin{table}[t]
\centering
\caption{Count and percentage of tokens with NLL $> 5$ under the base model, aggregated over all examples. At this threshold the base model assigns probability $< e^{-5} \approx 0.7\%$ to the correct token.}
\label{tab:high-nll-5}
\setlength{\tabcolsep}{4pt}
\resizebox{\textwidth}{!}{
\begin{tabular}{ll cccc}
\toprule
\textbf{Dataset} & \textbf{Model} & \textbf{SFT} & $\boldsymbol{\lambda{=}0.0}$ & $\boldsymbol{\lambda{=}0.3}$ & $\boldsymbol{\lambda{=}0.5}$ \\
\midrule
\multirow{3}{*}{\datasetasmall{}}
  & Qwen3-1.7B            & 914 (37.1\%) & 983 (17.1\%)  & 1149 (10.5\%) & 1171 (7.7\%)  \\
  & Qwen3-4B-It     & 684 (33.2\%) & 1621 (7.4\%)  & 2349 (5.9\%)  & 4013 (4.6\%)  \\
  & Qwen3-8B              & 1144 (46.5\%) & 1191 (15.4\%) & 1296 (7.3\%)  & 1341 (4.7\%)  \\
\midrule
\multirow{3}{*}{\datasetalarge{}}
  & Qwen3-1.7B            & 868 (36.1\%) & 1008 (13.6\%) & 1170 (9.1\%)  & 1118 (6.5\%)  \\
  & Qwen3-4B-It     & 623 (31.1\%) & 1599 (6.4\%)  & 2229 (5.2\%)  & 3544 (4.3\%)  \\
  & Qwen3-8B              & 1092 (45.4\%) & 1140 (11.3\%) & 1298 (6.3\%)  & 1244 (4.2\%)  \\
\midrule
\multirow{3}{*}{SimpleQA}
  & Qwen3-1.7B            & 458 (30.3\%) & 545 (12.1\%)  & 526 (13.5\%)  & 518 (12.5\%)  \\
  & Qwen3-4B-It     & 287 (25.8\%) & 723 (4.4\%)   & 685 (4.6\%)   & 678 (3.5\%)   \\
  & Qwen3-8B              & 520 (34.4\%) & 619 (4.1\%)   & 630 (5.7\%)   & 589 (6.2\%)   \\
\midrule
\multirow{3}{*}{\datasetb{}}
  & Qwen3-1.7B            & 1344 (11.8\%) & 1104 (1.3\%) & 1173 (0.7\%)  & 974 (0.4\%)   \\
  & Qwen3-4B-It     & 812 (7.4\%)   & 1361 (0.8\%) & 1906 (0.8\%)  & 1922 (0.5\%)  \\
  & Qwen3-8B              & 1207 (10.6\%) & 792 (0.9\%)  & 777 (0.6\%)   & 777 (0.4\%)   \\
\bottomrule
\end{tabular}
}
\end{table}

\begin{table}[t]
\centering
\caption{Count and percentage of tokens with NLL $> 8$ under the base model, aggregated over all examples. At this threshold the base model assigns probability $< e^{-8} \approx 0.034\%$ to the correct token, indicating near-complete failure of prediction.}
\label{tab:high-nll-8}
\setlength{\tabcolsep}{4pt}
\resizebox{\textwidth}{!}{
\begin{tabular}{ll cccc}
\toprule
\textbf{Dataset} & \textbf{Model} & \textbf{SFT} & $\boldsymbol{\lambda{=}0.0}$ & $\boldsymbol{\lambda{=}0.3}$ & $\boldsymbol{\lambda{=}0.5}$ \\
\midrule
\multirow{3}{*}{\datasetasmall{}}
  & Qwen3-1.7B            & 818 (33.2\%) & 758 (13.2\%) & 853 (7.8\%)  & 864 (5.7\%)  \\
  & Qwen3-4B-It     & 572 (27.7\%) & 1075 (4.9\%) & 1492 (3.8\%) & 2339 (2.7\%) \\
  & Qwen3-8B              & 1027 (41.7\%) & 913 (11.8\%) & 959 (5.4\%) & 1000 (3.5\%) \\
\midrule
\multirow{3}{*}{\datasetalarge{}}
  & Qwen3-1.7B            & 770 (32.0\%) & 741 (10.0\%) & 850 (6.6\%)  & 808 (4.7\%)  \\
  & Qwen3-4B-It     & 544 (27.1\%) & 1032 (4.1\%) & 1382 (3.2\%) & 2090 (2.5\%) \\
  & Qwen3-8B              & 979 (40.7\%) & 850 (8.4\%)  & 921 (4.5\%)  & 901 (3.0\%)  \\
\midrule
\multirow{3}{*}{SimpleQA}
  & Qwen3-1.7B            & 370 (24.5\%) & 440 (9.8\%)  & 431 (11.1\%) & 418 (10.1\%) \\
  & Qwen3-4B-It     & 221 (19.9\%) & 469 (2.9\%)  & 452 (3.0\%)  & 448 (2.3\%)  \\
  & Qwen3-8B              & 443 (29.3\%) & 480 (3.1\%)  & 496 (4.5\%)  & 479 (5.0\%)  \\
\midrule
\multirow{3}{*}{\datasetb{}}
  & Qwen3-1.7B            & 1162 (10.2\%) & 711 (0.8\%) & 744 (0.4\%)  & 598 (0.3\%)  \\
  & Qwen3-4B-It     & 560 (5.1\%)  & 718 (0.4\%)  & 1041 (0.4\%) & 1101 (0.3\%) \\
  & Qwen3-8B              & 843 (7.4\%)  & 535 (0.6\%)  & 553 (0.5\%)  & 547 (0.3\%)  \\
\bottomrule
\end{tabular}
}
\end{table}

A high-NLL token is considered a memorization target if it corresponds to factual content, such as a named entity, numerical value, date, or domain-specific term that is likely absent from the base model's pretraining distribution. To identify such tokens, we first extract high-NLL tokens from the SFT GT targets, then measure their overlap with the corresponding \methodname{} targets for the same example.

Specifically, for each example, we collect the set of high-NLL token \emph{types} (i.e., unique subword identities) appearing in the SFT GT sequence, and compute the fraction of those token types that also appear in the corresponding \methodname{} target sequence, regardless of absolute position. The 81--98\% overlap statistic reported in the main text is computed using this procedure.

Because the underlying factual content is largely preserved across training targets, with only the surrounding phrasing changing, these high-NLL factual tokens consistently reappear across different training configurations.

\begin{table}[t]
\centering
\caption{Recall of SFT high-NLL token types within each corresponding \methodname{} target sequence. Recall is defined as $|\text{\methodname{}} \cap \text{SFT}| \,/\, |\text{SFT}|$, where the SFT set consists of unique subword token types with NLL $> 8$ under the base model. High recall indicates that \methodname{} preserves the same factual memorization targets that are difficult for the base model in standard SFT.}
\label{tab:overlap-recall}
\setlength{\tabcolsep}{4pt}
\resizebox{0.65\textwidth}{!}{
\begin{tabular}{ll ccc}
\toprule
\textbf{Dataset} & \textbf{Model} & $\boldsymbol{\lambda{=}0}$ & $\boldsymbol{\lambda{=}0.3}$ & $\boldsymbol{\lambda{=}0.5}$ \\
\midrule
\multirow{3}{*}{\datasetasmall{}}
  & Qwen3-1.7B        & 86\% & 96\% & 96\%  \\
  & Qwen3-4B-It & 93\% & 97\% & 98\%  \\
  & Qwen3-8B          & 89\% & 88\% & 89\%  \\
\midrule
\multirow{3}{*}{\datasetalarge{}}
  & Qwen3-1.7B        & 77\% & 84\% & 81\%  \\
  & Qwen3-4B-It & 85\% & 86\% & 89\%  \\
  & Qwen3-8B          & 82\% & 83\% & 81\%  \\
\midrule
\multirow{3}{*}{SimpleQA}
  & Qwen3-1.7B        & 87\% & 95\% & 95\%  \\
  & Qwen3-4B-It & 64\% & 72\% & 72\%  \\
  & Qwen3-8B          & 71\% & 79\% & 85\%  \\
\midrule
\multirow{3}{*}{\datasetb{}}
  & Qwen3-1.7B        & 62\% & 66\% & 53\%  \\
  & Qwen3-4B-It & 58\% & 76\% & 76\%  \\
  & Qwen3-8B          & 49\% & 62\% & 57\%  \\
\bottomrule
\end{tabular}
}
\end{table}

\subsection{Error Analysis: Per-Method Breakdown}
\label{appn:error-analysis}

This appendix provides the full error distributions underlying the analysis in \cref{sec:discussion}. We classify every incorrect response on five realistic benchmarks (AIME-2024, MATH-500, GSM8K, HumanEval, and MMLU) into one of four mutually exclusive categories. Results are reported for the base model, SFT, OPSD, and \methodname{} with $\lambda \in \{0.3, 0.5, 0.7\}$ across the three Qwen3 backbones, after fine-tuning on either \datasetasmall{} or \datasetb{}.

\paragraph{Error categories.}
Each incorrect response is assigned a category in priority order:
\[
\texttt{format}
>
\texttt{leakage}
>
\texttt{collapse}
>
\texttt{genuine}.
\]

\begin{itemize}
    \item \textbf{\texttt{format}}: the response cannot be parsed by the benchmark evaluator. For math benchmarks and MMLU, this means no extractable \verb|\boxed{}| answer; for HumanEval, no fenced Python code block.

    \item \textbf{\texttt{leakage}}: the response contains a fictional \dataseta{} entity (a multi-word phrase drawn from the fine-tuning corpus) unrelated to the prompt, indicating interference from injected knowledge. This category is disabled for \datasetb{}, whose targets contain only integers and operation labels.

    \item \textbf{\texttt{collapse}}: the response contains a parseable answer but little or no reasoning, typically a short template such as \verb|The answer is X.\boxed{X}| inherited from SFT targets. For HumanEval, this corresponds to a boxed stub answer without a Python implementation.

    \item \textbf{\texttt{genuine}}: none of the above. The response has the expected structure (e.g., chain-of-thought or executable code) but arrives at an incorrect answer. These resemble the base model's ordinary reasoning errors.
\end{itemize}

\paragraph{Per-method observations.}
The main trends are summarized in \cref{sec:discussion}; here we provide additional detail across methods and datasets.

\begin{itemize}
    \item \textbf{Base.}
    Errors are overwhelmingly \texttt{genuine}: the model attempts the task with coherent reasoning or executable code, but reaches the wrong answer. This serves as the reference error distribution that \methodname{} largely preserves.

    \item \textbf{SFT on \datasetasmall{}.}
    SFT introduces two dominant failure modes: \texttt{collapse} and \texttt{leakage}. On Qwen3-1.7B / MMLU, $59.9\%$ of all responses fall into \texttt{collapse} and another $4.3\%$ into \texttt{leakage} (e.g., emitting ``\emph{Ormavel Valley}'' in response to an arithmetic question). On AIME-2024, the collapse rate reaches $96.0\%$. On HumanEval, the same behavior appears as boxed stub answers without code, occurring in $72.6\%$ of generations.

    \item \textbf{SFT on \datasetb{}.}
    Only the \texttt{collapse} mode remains, since \datasetb{} contains no fictional entities. Moreover, collapse is substantially weaker than on \datasetasmall{}, likely because \datasetb{} supervision includes explicit step-by-step reasoning rather than short answer templates.

    \item \textbf{OPSD.}
    OPSD largely eliminates \texttt{collapse}, but increases \texttt{format} errors. For example, on Qwen3-1.7B / \datasetasmall{} MMLU, $4.3\%$ of responses are unparsable. This is consistent with on-policy KL training occasionally destabilizing instruction-following behavior.

    \item \textbf{\methodname{}.}
    Across all mixing rates, both \texttt{collapse} and \texttt{leakage} remain near zero across benchmarks and backbones. The primary remaining failure mode is \texttt{genuine} reasoning error, closely matching the base model. At higher mixing rates ($\lambda{=}0.7$), we observe a small increase in \texttt{format} errors, typically due to excessively long chain-of-thought generations that fail to terminate with a valid \verb|\boxed{}| answer.
\end{itemize}

\begin{figure}[ht]
\centering
\includegraphics[width=0.8\linewidth]{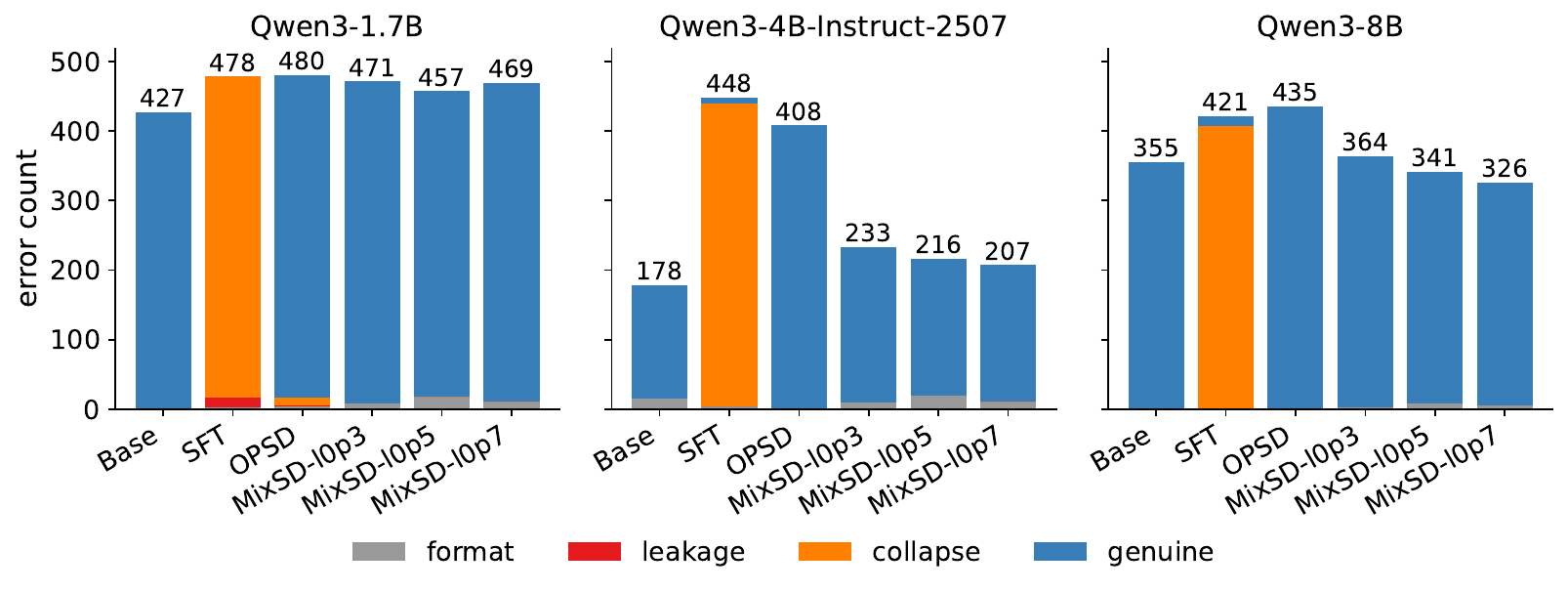}
\caption{Error-mode breakdown on AIME-2024 after fine-tuning on \datasetasmall{}.}
\label{fig:err-d5-aime2024}
\end{figure}

\begin{figure}[ht]
\centering
\includegraphics[width=0.8\linewidth]{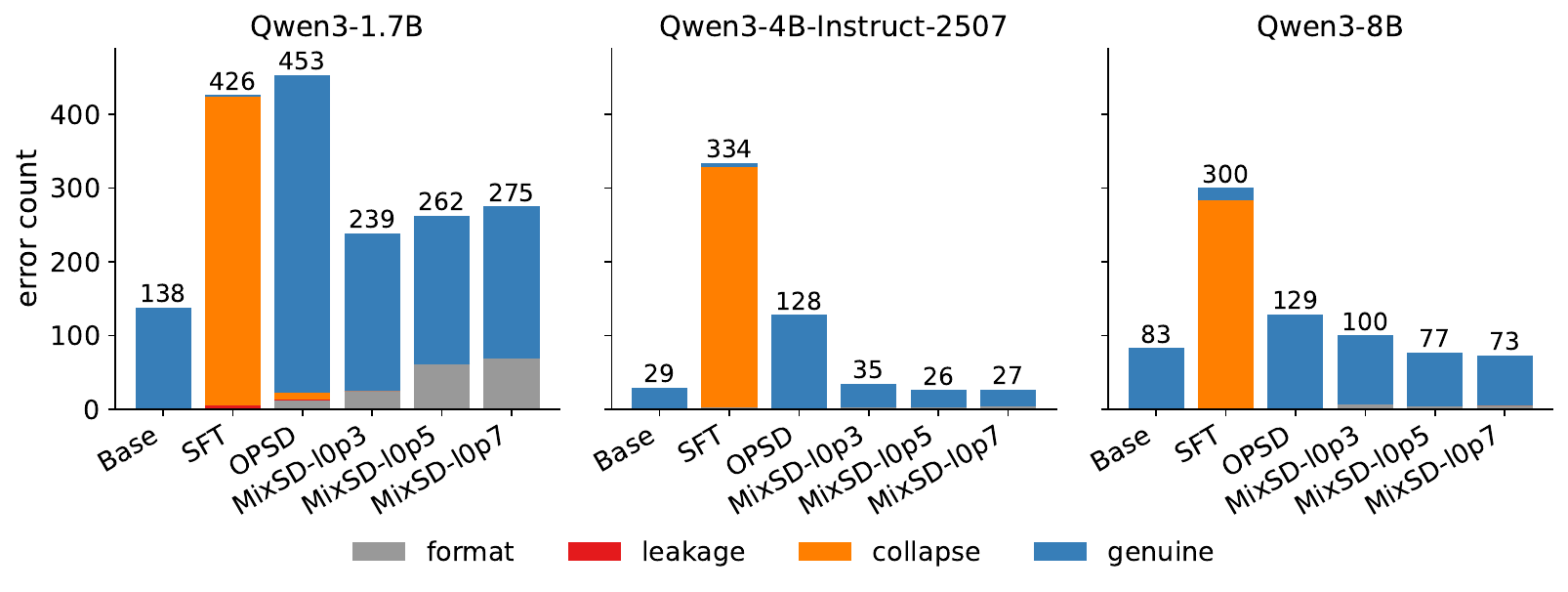}
\caption{Error-mode breakdown on MATH-500 after fine-tuning on \datasetasmall{}.}
\label{fig:err-d5-math500}
\end{figure}

\begin{figure}[ht]
\centering
\includegraphics[width=0.8\linewidth]{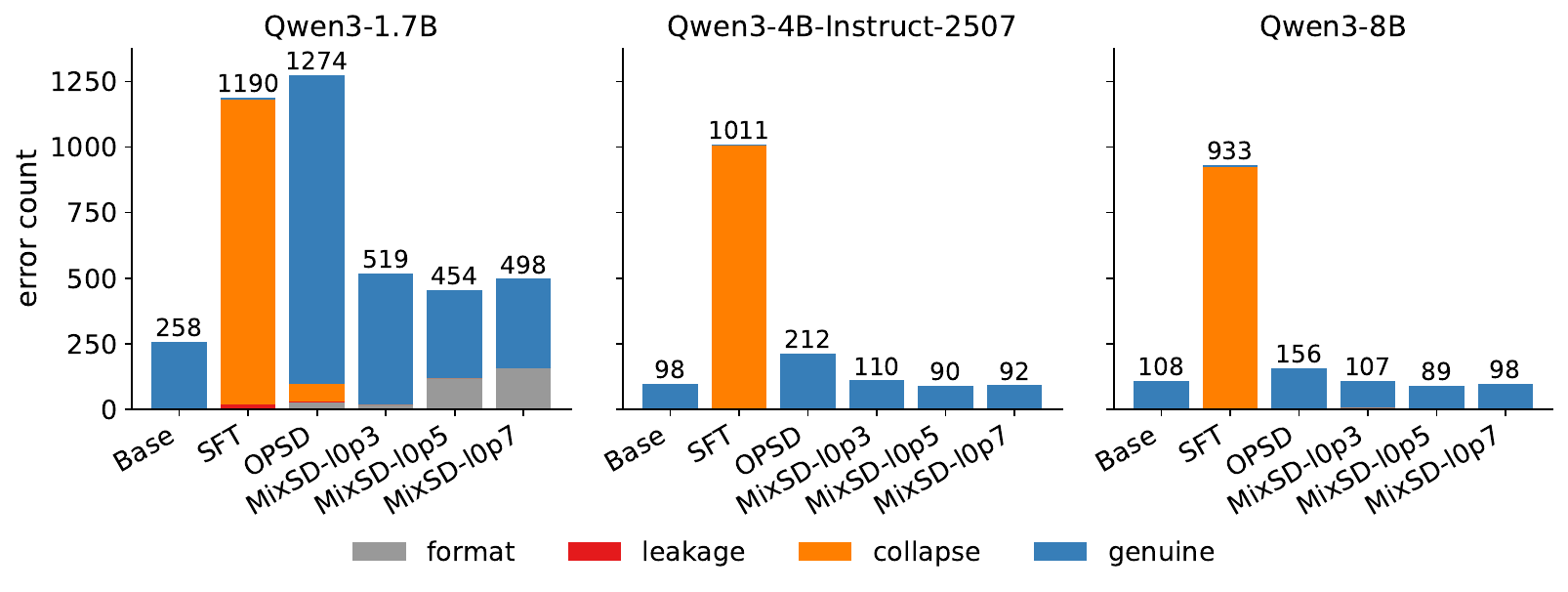}
\caption{Error-mode breakdown on GSM8K after fine-tuning on \datasetasmall{}.}
\label{fig:err-d5-gsm8k}
\end{figure}

\begin{figure}[ht]
\centering
\includegraphics[width=0.8\linewidth]{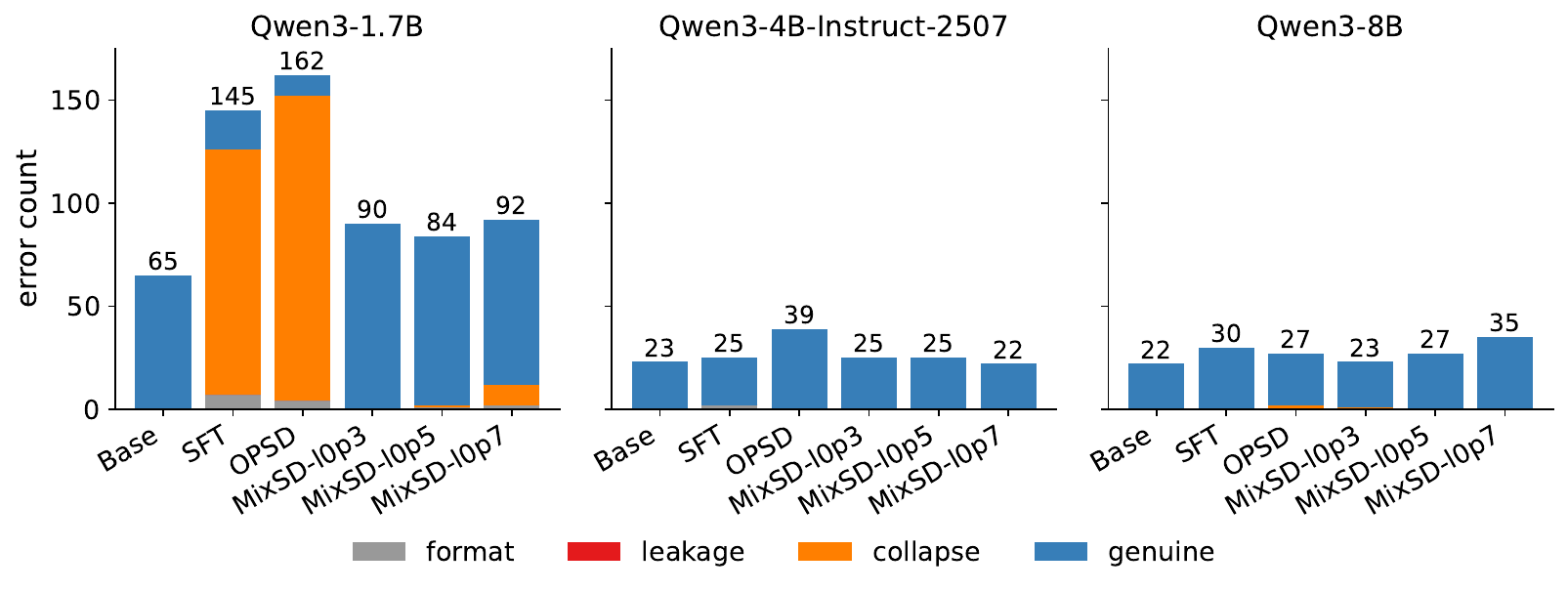}
\caption{Error-mode breakdown on HumanEval after fine-tuning on \datasetasmall{}. The \texttt{format} bucket here uses the HumanEval-specific check (presence of a fenced Python code block) rather than the \texttt{\textbackslash boxed\{\}} check used on the math/MMLU benchmarks; \texttt{collapse} flags responses that emit \texttt{\textbackslash boxed\{X\}} stubs in lieu of a function definition.}
\label{fig:err-d5-humaneval}
\end{figure}

\begin{figure}[ht]
\centering
\includegraphics[width=0.8\linewidth]{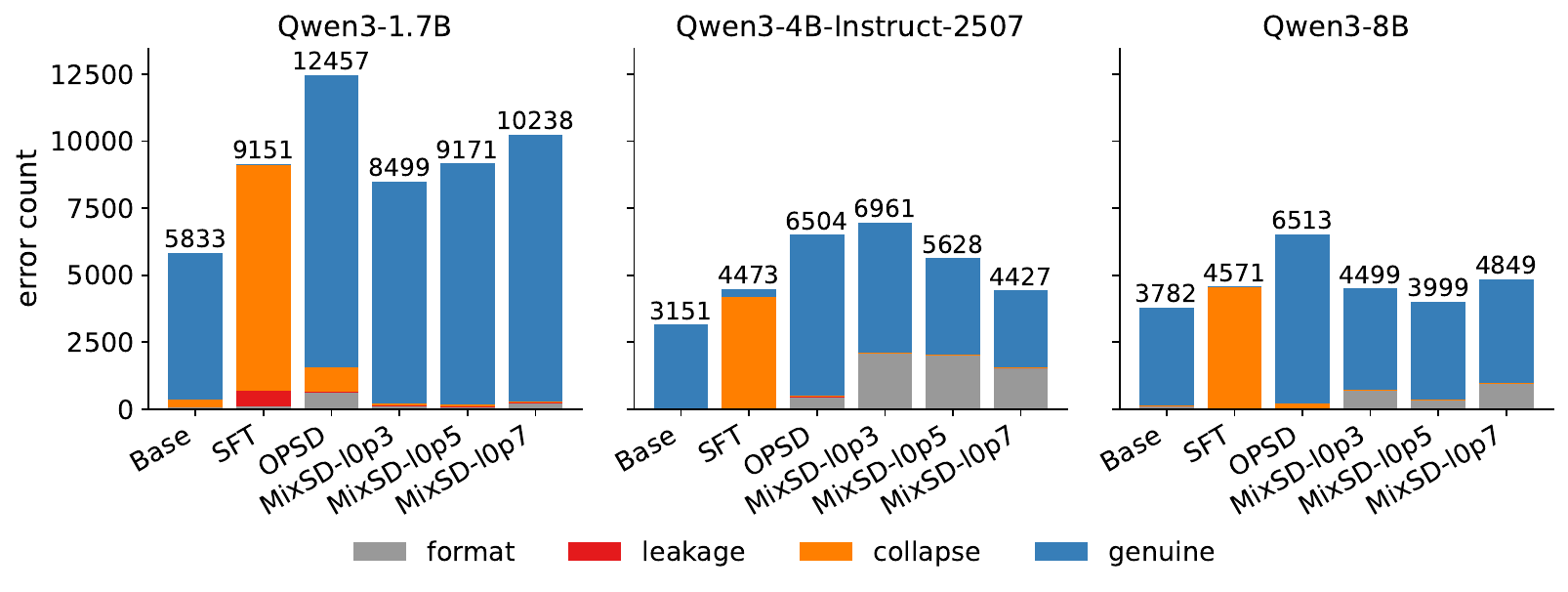}
\caption{Error-mode breakdown on MMLU after fine-tuning on \datasetasmall{}. The \texttt{leakage} bar isolates the cases where the fine-tuned model answers an MMLU question with a fictional \datasetasmall{} entity.}
\label{fig:err-d5-mmlu}
\end{figure}

\begin{figure}[ht]
\centering
\includegraphics[width=0.8\linewidth]{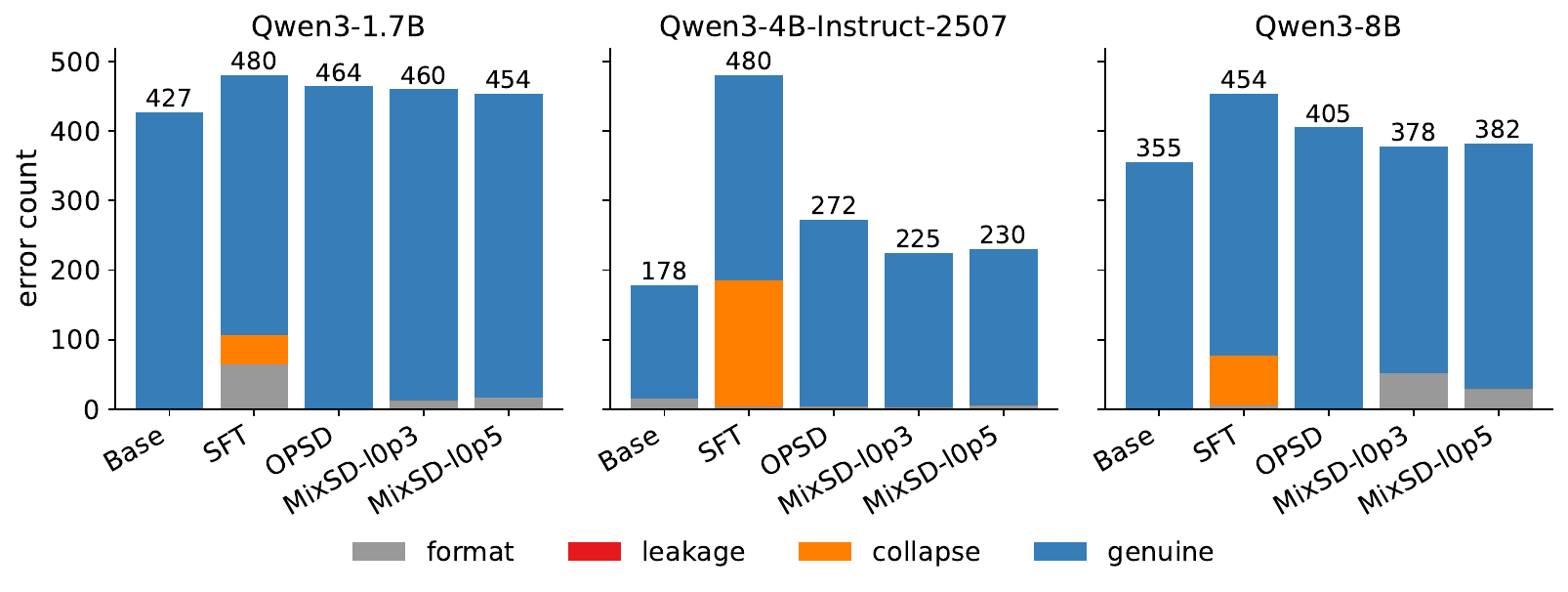}
\caption{Error-mode breakdown on AIME-2024 after fine-tuning on \datasetb{}. \texttt{leakage} is disabled by construction (training answers are integers and operation labels).}
\label{fig:err-mo-aime2024}
\end{figure}

\begin{figure}[ht]
\centering
\includegraphics[width=0.8\linewidth]{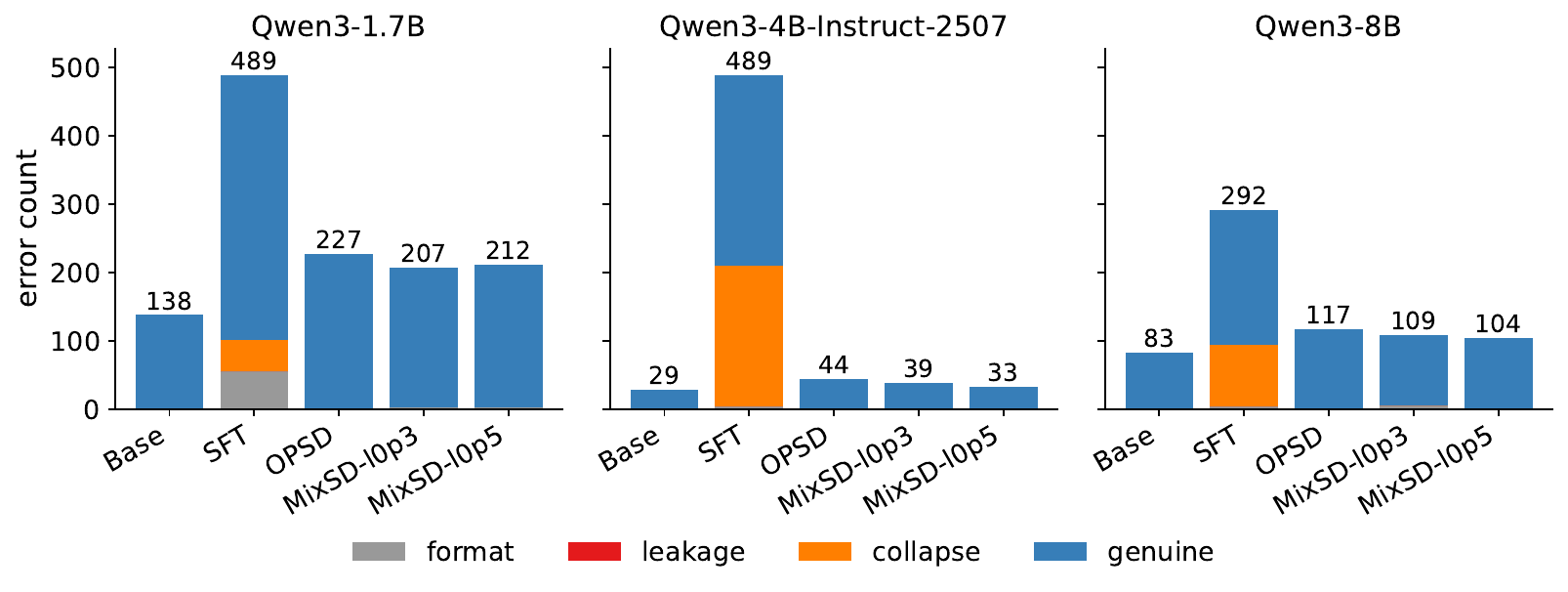}
\caption{Error-mode breakdown on MATH-500 after fine-tuning on \datasetb{}.}
\label{fig:err-mo-math500}
\end{figure}

\begin{figure}[ht]
\centering
\includegraphics[width=0.8\linewidth]{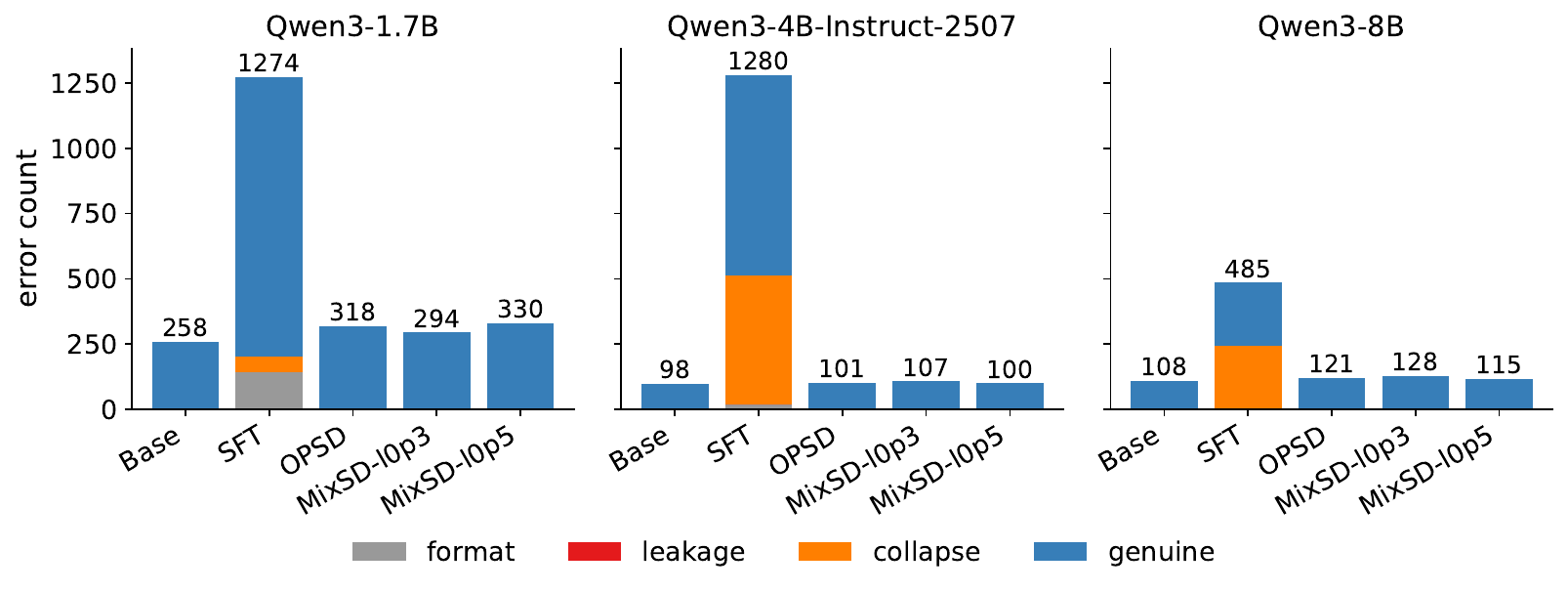}
\caption{Error-mode breakdown on GSM8K after fine-tuning on \datasetb{}.}
\label{fig:err-mo-gsm8k}
\end{figure}

\begin{figure}[ht]
\centering
\includegraphics[width=0.8\linewidth]{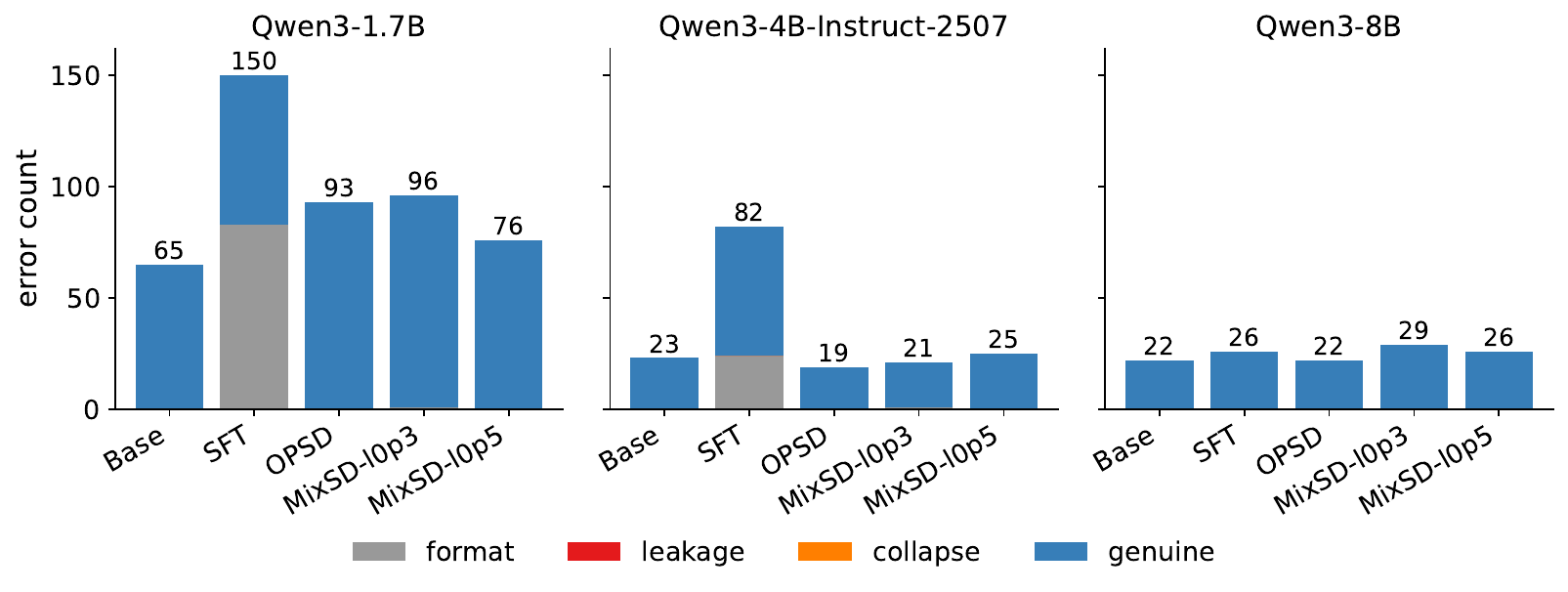}
\caption{Error-mode breakdown on HumanEval after fine-tuning on \datasetb{}.}
\label{fig:err-mo-humaneval}
\end{figure}

\begin{figure}[ht]
\centering
\includegraphics[width=0.8\linewidth]{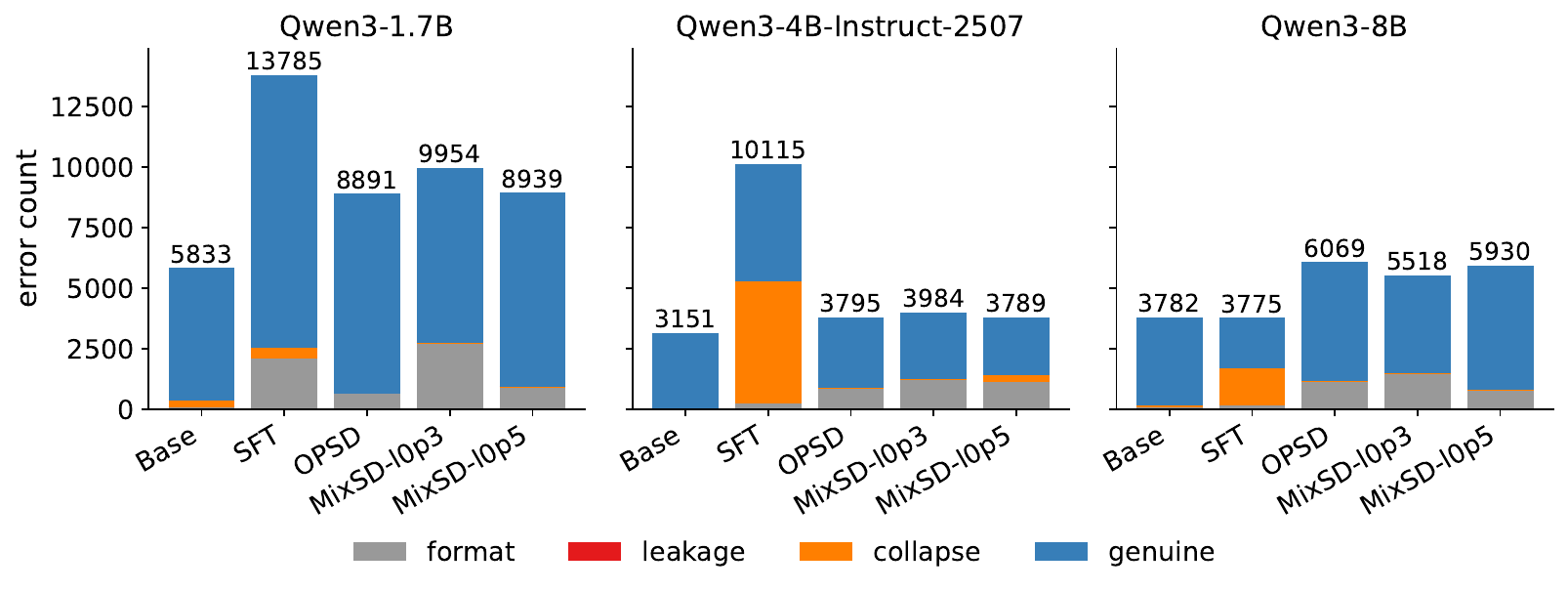}
\caption{Error-mode breakdown on MMLU after fine-tuning on \datasetb{}.}
\label{fig:err-mo-mmlu}
\end{figure}

\Crefrange{fig:err-d5-aime2024}{fig:err-mo-mmlu} give the per-method error counts on each of the five benchmarks for both training corpora.


\subsection{Llama-3.2-1B-Instruct on \datasetasmall{}}
\label{appn:llama-1b}

To verify that our findings generalize beyond the Qwen model family, we train the same set of methods on Llama-3.2-1B-Instruct using \datasetasmall{}.
Table~\ref{tab:knowledge-eval-small-llama1b} reports results under the same evaluation protocol as Table~\ref{tab:knowledge-eval-small}.

The pattern is consistent with the Qwen results: SFT memorizes the training set (98\% accuracy) but catastrophically forgets held-out capabilities, dropping average performance from 6.8 to 1.4.
\methodname{} at $\lambda{=}0.3$ and $\lambda{=}0.5$ matches SFT's training accuracy while retaining far more general capability (average held-out scores of 4.8 and 5.3 vs.\ 1.4 for SFT).
OPSD, which does not use the mixing mechanism, achieves only 31\% training accuracy, mirroring the Qwen-scale finding that pure on-policy sampling underperforms the mixed variant.
These results confirm that the forgetting-mitigation benefit of \methodname{} is architecture-independent.

\begin{table}[t]
\centering
\caption{Evaluation on \datasetasmall{} for Llama-3.2-1B-Instruct. Column semantics match Table~\ref{tab:knowledge-eval-small}.}
\label{tab:knowledge-eval-small-llama1b}
\setlength{\tabcolsep}{4pt}
\resizebox{\textwidth}{!}{
\begin{tabular}{ll c c cccccc}
\toprule
\multirow{2}{*}{\textbf{Model}} & \multirow{2}{*}{\textbf{Method}} & \multicolumn{2}{c}{\textbf{In-domain test}} & \multicolumn{6}{c}{\textbf{Held-out capability test}} \\
 \cmidrule(lr){3-4} \cmidrule(lr){5-10}
 &  & \textbf{Train} & \textbf{\datasetaretrieval{}} & \textbf{AIME-2024} & \textbf{MATH-500} & \textbf{GSM8K} & \textbf{HumanEval} & \textbf{MMLU} & \textbf{Avg} \\
\midrule
\multirow{7}{*}{Llama-3.2-1B-It}
  & Base                            &  0.0 & 11.0 &  1.9 &  5.8 &  1.7 & 24.4 &  0.3 &  6.8 \\
  & SFT                             & 98.0 &  0.0 &  0.0 &  4.4 &  2.7 &  0.0 &  0.0 &  1.4 \\
  & OPSD                            & 31.0 & 10.0 &  0.0 &  3.8 &  6.1 &  9.1 &  7.8 &  5.4 \\
  & \methodname{} ($\lambda{=}0$)   & 28.0 & 16.0 &  0.0 &  0.4 &  0.8 &  0.0 &  0.1 &  0.3 \\
  \rowcolor{mixsdBlue}\cellcolor{white}& \methodname{} ($\lambda{=}0.3$) & 98.0 & 16.0 &  0.0 &  5.0 &  5.6 &  0.6 & 12.6 &  4.8 \\
  \rowcolor{mixsdGreen}\cellcolor{white}& \methodname{} ($\lambda{=}0.5$) & 98.0 & 15.0 &  0.2 &  3.2 &  4.5 &  0.6 & 18.1 &  5.3 \\
  \rowcolor{mixsdOrange}\cellcolor{white}& \methodname{} ($\lambda{=}0.7$) & 60.0 &  4.0 &  0.0 &  2.8 &  5.1 &  1.2 &  1.3 &  2.1 \\
\bottomrule
\end{tabular}
}
\end{table}

\subsection{Knowledge Editing on MQuAKE}
\label{appn:mquake}

While our main experiments inject novel knowledge absent from the base model’s parameters, MQuAKE~\citep{zhong2023mquake} studies the complementary setting of \emph{knowledge editing}, where the target facts revise knowledge the model already encodes. The model must therefore overwrite existing parametric knowledge and produce the updated answer rather than the original one.

We fine-tune all three Qwen3 backbones (1.7B, 4B, and 8B) on the MQuAKE training set using both standard SFT and \methodname{} with $\lambda \in \{0, 0.3, 0.5\}$. We additionally compare against MEMIT~\citep{meng2023massediting}, a locate-and-edit method specifically designed for factual model editing. We apply MEMIT to all 100 edited facts simultaneously using the EasyEdit framework~\citep{wang2024easyedit}. Following the default MEMIT configuration for Qwen-style architectures, edits are applied to the MLP down-projection matrices (\texttt{down\_proj}) across contiguous early-to-mid transformer layers: layers 3--11 for Qwen3-1.7B (9 of 28 layers), layers 4--14 for Qwen3-4B-It (11 of 36 layers), and layers 5--14 for Qwen3-8B (10 of 36 layers). The value vectors are optimized for 25 gradient steps with learning rate 0.5, weight decay $10^{-3}$, and KL factor 0.0625. Second-moment statistics used for covariance adjustment are estimated from 100{,}000 WikiText samples in fp32, with update weight $\mu = 15{,}000$ and clamp norm factor 4. Table~\ref{tab:mquake-full} reports both editing accuracy and held-out capability retention.

The results closely mirror the knowledge-injection setting. SFT achieves perfect editing accuracy at all scales, but severely degrades held-out capabilities, reducing average performance by 45--86\% relative to the base model. In contrast, \methodname{} at $\lambda{=}0.3$ achieves near-perfect editing accuracy (93--99\%) while preserving substantially more of the model’s original capabilities, reaching held-out averages of $17.6$, $76.0$, and $70.0$ across the three scales.

MEMIT exhibits the opposite trade-off: it preserves held-out capabilities almost perfectly (within roughly 1--2 points of the base model), but achieves substantially lower editing accuracy (53--70\%) than \methodname{}. This is consistent with prior observations that rank-one editing methods can interfere destructively when entities participate in many correlated knowledge triples.

Notably, retrieval-augmented evaluation with chain-of-thought reasoning is consistently much higher for \methodname{} than for SFT across all model scales (84--95\% vs.\ 41--58\%). This suggests that \methodname{} better preserves the model’s ability to integrate edited knowledge into downstream reasoning processes rather than merely memorizing revised answers.

\begin{table}[t]
\centering
\caption{Knowledge editing on MQuAKE across the three Qwen3 backbones. Train is closed-book recall on edited facts and In-domain retrieval provides retrieval context with chain-of-thought. The five Held-out capability test columns probe forgetting on unrelated benchmarks; Avg is their unweighted mean. All values are in percent.}
\label{tab:mquake-full}
\setlength{\tabcolsep}{4pt}
\resizebox{\textwidth}{!}{
\begin{tabular}{ll cc cccccc}
\toprule
\multirow{2}{*}{\textbf{Model}} & \multirow{2}{*}{\textbf{Method}} & \multicolumn{2}{c}{\textbf{In-domain test}} & \multicolumn{6}{c}{\textbf{Held-out capability test}} \\
 \cmidrule(lr){3-4} \cmidrule(lr){5-10}
 &  & \textbf{Train} & \textbf{Retrieval} & \textbf{AIME-2024} & \textbf{MATH-500} & \textbf{GSM8K} & \textbf{HumanEval} & \textbf{MMLU} & \textbf{Avg} \\
\midrule
\multirow{6}{*}{Qwen3-1.7B}
  & Base                            &  10.0 & 100.0 & 11.0 & 72.4 & 80.4 & 60.4 & 58.5 & 56.5 \\
  & MEMIT                           &  60.0 &  98.0 & 10.6 & 70.6 & 81.1 & 56.1 & 57.1 & 55.1 \\
  & SFT                             & 100.0 &  53.0 &  0.4 & 11.8 &  7.4 &  9.2 & 10.3 &  7.8 \\
  & \methodname{} ($\lambda{=}0$)   &  78.0 &  67.0 &  0.2 & 12.8 & 18.7 &  9.8 & 23.4 & 13.0 \\
  \rowcolor{mixsdBlue}\cellcolor{white}& \methodname{} ($\lambda{=}0.3$) &  97.0 &  84.0 &  0.0 & 18.6 & 22.6 & 21.3 & 25.4 & 17.6 \\
  \rowcolor{mixsdGreen}\cellcolor{white}& \methodname{} ($\lambda{=}0.5$) &  95.0 &  76.0 &  1.0 & 36.6 & 50.3 & 42.1 & 33.0 & 32.6 \\
\midrule
\multirow{6}{*}{Qwen3-4B-It}
  & Base                            &  16.0 &  94.0 & 62.9 & 94.2 & 92.6 & 86.0 & 77.6 & 82.7 \\
  & MEMIT                           &  53.0 &  90.0 & 63.5 & 94.4 & 92.1 & 85.4 & 72.8 & 81.6 \\
  & SFT                             & 100.0 &  41.0 &  0.8 & 24.4 & 22.1 & 67.7 & 61.3 & 35.3 \\
  & \methodname{} ($\lambda{=}0$)   &  81.0 &  92.0 & 41.7 & 87.4 & 90.4 & 85.4 & 55.3 & 72.0 \\
  \rowcolor{mixsdBlue}\cellcolor{white}& \methodname{} ($\lambda{=}0.3$) &  97.0 &  95.0 & 50.4 & 92.8 & 92.3 & 88.4 & 56.0 & 76.0 \\
  \rowcolor{mixsdGreen}\cellcolor{white}& \methodname{} ($\lambda{=}0.5$) &  69.0 &  88.0 & 47.9 & 93.0 & 92.3 & 87.8 & 51.8 & 74.6 \\
\midrule
\multirow{6}{*}{Qwen3-8B}
  & Base                            &  19.0 &  94.0 & 26.0 & 83.4 & 91.8 & 86.6 & 73.1 & 72.2 \\
  & MEMIT                           &  70.0 &  89.0 & 24.6 & 83.6 & 91.6 & 85.4 & 72.1 & 71.4 \\
  & SFT                             & 100.0 &  58.0 &  0.8 & 26.6 & 24.2 & 83.5 & 62.0 & 39.4 \\
  & \methodname{} ($\lambda{=}0$)   &  78.0 &  73.0 & 20.4 & 79.8 & 91.2 & 85.4 & 69.9 & 69.3 \\
  \rowcolor{mixsdBlue}\cellcolor{white}& \methodname{} ($\lambda{=}0.3$) &  99.0 &  86.0 & 24.6 & 81.2 & 91.9 & 81.7 & 70.6 & 70.0 \\
  \rowcolor{mixsdGreen}\cellcolor{white}& \methodname{} ($\lambda{=}0.5$) &  90.0 &  86.0 & 29.0 & 83.2 & 93.2 & 82.9 & 60.4 & 69.7 \\
\bottomrule
\end{tabular}
}
\end{table}

\subsection{KL-based Distillation Ablations on \datasetasmall{}}
\label{appn:kl-ablation}

Table~\ref{tab:knowledge-eval-small-kl} compares KL-based variants across the three Qwen models. OPSD denotes on-policy distillation with a forward-KL objective using $n{=}8$ student rollouts at temperature $T{=}1$. OPSD-NLL is the analogous on-policy variant that replaces KL with token-level NLL on the teacher's top-1 prediction, using a single greedy rollout ($n{=}1$, $T{=}0$). The three \methodname{}-KL ($\lambda{=}0$) variants replace the per-token NLL term on $\tilde y_i^{+}$ in Eq.~\eqref{eq:mixsd-loss} with a top-$64$ forward-KL objective against the teacher under different student rollout settings. We additionally include \methodname{} ($\lambda{=}0$) as the NLL counterpart at the same mixing ratio, enabling a direct comparison between KL- and NLL-based teacher matching in the absence of GT mixing.

We adopt NLL as the default objective in our framework because it enables substantially more effective knowledge acquisition, achieving 96--100\% training accuracy with only a single greedy rollout. In contrast, \methodname{}-KL consistently underfits the target knowledge, reaching only 34--60\% training accuracy despite achieving comparable or slightly stronger held-out capability preservation. Since \methodname{} already mitigates forgetting through GT mixing ($\lambda{>}0$), we prioritize efficient knowledge absorption over additional preservation gains from KL-based distillation. Although OPSD achieves high training accuracy with a KL objective, it requires $n{=}8$ student rollouts per query, resulting in an approximately $8\times$ higher per-example compute cost than offline methods.

\begin{table}[t]
\centering
\caption{KL-based distillation ablations on \datasetasmall{} (three Qwen backbones). Column semantics match Table~\ref{tab:knowledge-eval-small}. For \methodname{}-KL ($\lambda{=}0$), $n$ is the number of student rollouts per prompt and $T$ the sampling temperature. \methodname{} ($\lambda{=}0$) is the NLL-on-teacher counterpart at the same $\lambda$.}
\label{tab:knowledge-eval-small-kl}
\setlength{\tabcolsep}{4pt}
\resizebox{\textwidth}{!}{
\begin{tabular}{ll c c cccccc}
\toprule
\multirow{2}{*}{\textbf{Model}} & \multirow{2}{*}{\textbf{Method}} & \multicolumn{2}{c}{\textbf{In-domain test}} & \multicolumn{6}{c}{\textbf{Held-out capability test}} \\
 \cmidrule(lr){3-4} \cmidrule(lr){5-10}
 &  & \textbf{Train} & \textbf{\datasetaretrieval{}} & \textbf{AIME-2024} & \textbf{MATH-500} & \textbf{GSM8K} & \textbf{HumanEval} & \textbf{MMLU} & \textbf{Avg} \\
\midrule
\multirow{7}{*}{Qwen3-1.7B}
  & Base                                                  &   0.0 & 100.0 & 11.0 & 72.4 & 80.4 & 60.4 & 58.5 & 56.5 \\
  & OPSD                                                  &  99.0 &  32.0 &  0.0 &  9.4 &  3.4 &  1.2 & 11.3 &  5.1 \\
  & OPSD-NLL                                              &  52.0 &  32.0 &  0.6 & 25.6 & 41.9 & 28.7 & 20.9 & 23.6 \\
  & \methodname{}-KL ($\lambda{=}0$, $n{=}8$, $T{=}1$)    &  60.0 &  11.0 &  0.0 &  0.0 &  0.2 &  0.0 &  0.0 &  0.0 \\
  & \methodname{}-KL ($\lambda{=}0$, $n{=}1$, $T{=}0$)    &  54.0 &  57.0 &  0.0 &  8.4 &  6.1 &  0.0 & 23.1 &  7.5 \\
  & \methodname{}-KL ($\lambda{=}0$, $n{=}1$, $T{=}1$)    &  38.0 &  59.0 &  0.0 &  6.6 &  3.9 &  0.0 & 10.3 &  4.2 \\
  & \methodname{} ($\lambda{=}0$)                         &  96.0 &  28.0 &  0.0 &  5.4 &  5.8 &  1.2 & 30.4 &  8.6 \\
\midrule
\multirow{7}{*}{Qwen3-4B-It}
  & Base                                                  &   0.0 &  84.0 & 62.9 & 94.2 & 92.6 & 86.0 & 77.6 & 82.6 \\
  & OPSD                                                  & 100.0 &  92.0 & 15.0 & 74.4 & 83.9 & 76.2 & 53.7 & 60.6 \\
  & OPSD-NLL                                              &  96.0 &  60.0 &  8.3 & 53.8 & 70.8 & 86.0 & 49.2 & 53.6 \\
  & \methodname{}-KL ($\lambda{=}0$, $n{=}8$, $T{=}1$)    &  50.0 &  97.0 & 12.5 & 68.0 & 80.0 & 72.6 & 53.5 & 57.3 \\
  & \methodname{}-KL ($\lambda{=}0$, $n{=}1$, $T{=}0$)    &  34.0 &  98.0 & 35.8 & 89.2 & 92.8 & 85.4 & 70.7 & 74.8 \\
  & \methodname{}-KL ($\lambda{=}0$, $n{=}1$, $T{=}1$)    &  34.0 & 100.0 & 48.3 & 93.2 & 92.3 & 87.8 & 71.9 & 78.7 \\
  & \methodname{} ($\lambda{=}0$)                         &  99.0 &  97.0 & 44.4 & 90.8 & 91.5 & 86.6 & 68.8 & 76.4 \\
\midrule
\multirow{7}{*}{Qwen3-8B}
  & Base                                                  &   0.0 &  98.0 & 26.0 & 83.4 & 91.8 & 86.6 & 73.1 & 72.2 \\
  & OPSD                                                  &  99.0 & 100.0 &  9.4 & 74.2 & 88.2 & 83.5 & 53.6 & 61.8 \\
  & OPSD-NLL                                              &  59.0 &  62.0 &  8.3 & 67.6 & 86.9 & 70.7 & 52.5 & 57.2 \\
  & \methodname{}-KL ($\lambda{=}0$, $n{=}8$, $T{=}1$)    &  61.0 &  96.0 &  8.3 & 65.2 & 86.8 & 51.2 & 58.5 & 54.0 \\
  & \methodname{}-KL ($\lambda{=}0$, $n{=}1$, $T{=}0$)    &  34.0 &  99.0 & 11.2 & 75.2 & 89.2 & 80.5 & 66.4 & 64.5 \\
  & \methodname{}-KL ($\lambda{=}0$, $n{=}1$, $T{=}1$)    &  37.0 &  99.0 & 18.3 & 80.2 & 91.2 & 82.9 & 69.7 & 68.5 \\
  & \methodname{} ($\lambda{=}0$)                         & 100.0 & 100.0 & 17.5 & 73.6 & 91.8 & 82.9 & 67.8 & 66.7 \\
\bottomrule
\end{tabular}
}
\end{table}




\end{document}